\documentclass{article} % For LaTeX2e
\usepackage{iclr2022_conference,times}

\usepackage{amsmath,amsfonts,bm}

\def\eqref#1{equation~\ref{#1}}
\def\1{\bm{1}}

\def\vb{{\bm{b}}}

\def\vh{{\bm{h}}}

\def\vx{{\bm{x}}}

\def\mC{{\bm{C}}}

\def\mK{{\bm{K}}}

\def\mP{{\bm{P}}}
\def\mQ{{\bm{Q}}}

\def\mV{{\bm{V}}}
\def\mW{{\bm{W}}}

\DeclareMathAlphabet{\mathsfit}{\encodingdefault}{\sfdefault}{m}{sl}
\SetMathAlphabet{\mathsfit}{bold}{\encodingdefault}{\sfdefault}{bx}{n}

\def\sR{{\mathbb{R}}}

\usepackage{hyperref}
\usepackage{url}

\usepackage{amssymb}
\usepackage{amsbsy}
\usepackage{amsmath}
\usepackage{amsthm}
\usepackage{amsfonts}
\usepackage{latexsym}
\usepackage{subcaption}
\usepackage{wrapfig}

\usepackage{graphicx}
\usepackage{tabularx}
\usepackage{booktabs}
\usepackage{ulem}
\usepackage[font=small]{caption}
\usepackage{algorithm}
\usepackage{algorithmicx, algpseudocode}
\usepackage{mathtools}
\usepackage{arydshln}
\usepackage{titlesec}

\titlespacing{\paragraph}{%
  0pt}{%              left margin
  0.01 \baselineskip}{% space before (vertical)
  1em}% 

\title{Towards a Unified View of \\ Parameter-Efficient Transfer Learning}

\author{Junxian He\thanks{Equal Contribution. Order determined by random dice rolling.}\\ 
Carnegie Mellon University\\
\texttt{junxianh@cs.cmu.edu}
\And
Chunting Zhou$^*$ \\
Carnegie Mellon University\\
\texttt{chuntinz@cs.cmu.edu}
\And
Xuezhe Ma \\
University of Southern California\\
\texttt{xuezhema@isi.edu}
\And
Taylor Berg-Kirkpatrick \\
UC San Diego\\
\texttt{tberg@eng.ucsd.edu} 
\And
Graham Neubig \\
Carnegie Mellon University\\
\texttt{gneubig@cs.cmu.edu}
}

\newcommand{\wq}{\mW_q}
\newcommand{\wk}{\mW_k}
\newcommand{\wv}{\mW_v}
\newcommand{\wo}{\mW_o}
\newcommand{\wdd}{\mW_{\text{down}}}
\newcommand{\wu}{\mW_{\text{up}}}
\newcommand{\tc}{\text{concat}}

\newcommand{\pk}{\mP_k}
\newcommand{\pv}{\mP_v}
\newcommand{\dvh}{\Delta \vh}

\iclrfinalcopy % Uncomment for camera-ready version, but NOT for submission.
\begin{document}

\maketitle

\begin{abstract}
Fine-tuning large pretrained language models on downstream tasks has become the de-facto learning paradigm in NLP. However, conventional approaches fine-tune all the parameters of the pretrained model, which becomes prohibitive as the model size and the number of tasks grow. 
Recent work has proposed a variety of parameter-efficient transfer learning methods that only fine-tune a small number of (extra) parameters to attain strong performance. 
% While these works use different architectural modifications and study different downstream tasks, they come to the same conclusion: various approaches to efficient fine-tuning can achieve comparable performance to fine-tuning all model parameters. 
While effective, the critical ingredients for success and the connections among the various methods are poorly understood.
In this paper, we break down the design of state-of-the-art parameter-efficient transfer learning methods and present a unified framework that establishes connections between them.
Specifically, we re-frame them as modifications to specific hidden states in pretrained models, and define a set of design dimensions along which different methods vary, such as the function to compute the modification and the position to apply the modification.
% The unified view allows us to identify critical design choices and transfer techniques across approaches which yields new variants.
Through comprehensive empirical studies across machine translation, text summarization, language understanding, and text classification benchmarks, we utilize the unified view to identify important design choices in previous methods.
% showing that existing approaches have various advantages and limitations in different settings.
Furthermore, our unified framework enables the transfer of design elements across different approaches, and as a result we are able to instantiate new parameter-efficient fine-tuning methods that tune less parameters than previous methods while being more effective, achieving comparable results to fine-tuning all parameters on all four tasks.\footnote{
Code is available at \href{https://github.com/jxhe/unify-parameter-efficient-tuning}{https://github.com/jxhe/unify-parameter-efficient-tuning}.
}
% At a higher level, we hope our work can provide insights and guidance for future research to develop or deploy parameter-efficient transfer learning approaches.

\end{abstract}
%!TEX root=./iclr2022_conference.tex
\section{Introduction}

% \gn{We should come up with a name for the proposed method and use it throughout the figures and tables.}

% \gn{The term ``transfer'' was bugging me a little bit as I read through the paper and I finally came up with something I think I came up with terms that I like better: ``mix-and-match'' and ``plug in''. ``mix-and-match'' corresponds to picking your favorite elements of each existing work, and ``plug in'' is what you were previously calling ``transfer'' (we ``plug in the multi-head component of prefix tuning into adapters''). We could even call the proposed method a ``mix-and-match adapter'' (MAM-Adapt) or something like that if we wanted to. What do you think?}

% \gn{Reading all the way to the end, I think this is a really nice paper, but I think (1) we should try a little harder to sell our main and exciting results, and (2) try to explain the intuition behind them a little better, ideally also with analysis if possible to do by the deadline -- right now due to a lack of explanation (in some places) things still may seem a little bit mysterious to some readers.}

% \gn{What do you think about making parameter matrices not bold? Currently there's a ton of bold in the equations which makes it a little noisy, and I think it's probably sufficient to just express matrices through capitalization.}

Transfer learning from pre-trained language models (PLMs) is now the prevalent paradigm in natural language processing, yielding strong performance on many tasks~\citep{peters2018deep,devlin2019bert,qiu2020pre}. The most common way to adapt general-purpose PLMs to downstream tasks is to fine-tune all the model parameters (\emph{full fine-tuning}). However, this results in a separate copy of fine-tuned model parameters for each task, which is prohibitively expensive when serving models that perform a large number of tasks. This issue is particularly salient with the ever-increasing size of PLMs, which now range from hundreds of millions~\citep{radford2019language,lewis-etal-2020-bart} to hundreds of billions~\citep{brown2020language} or even trillions of parameters~\citep{fedus2021switch}.

\begin{figure}[!t]
    % \centering
    \begin{minipage}{0.49\textwidth}
    \begin{figure}[H]
        \centering
        \includegraphics[width=1\textwidth]{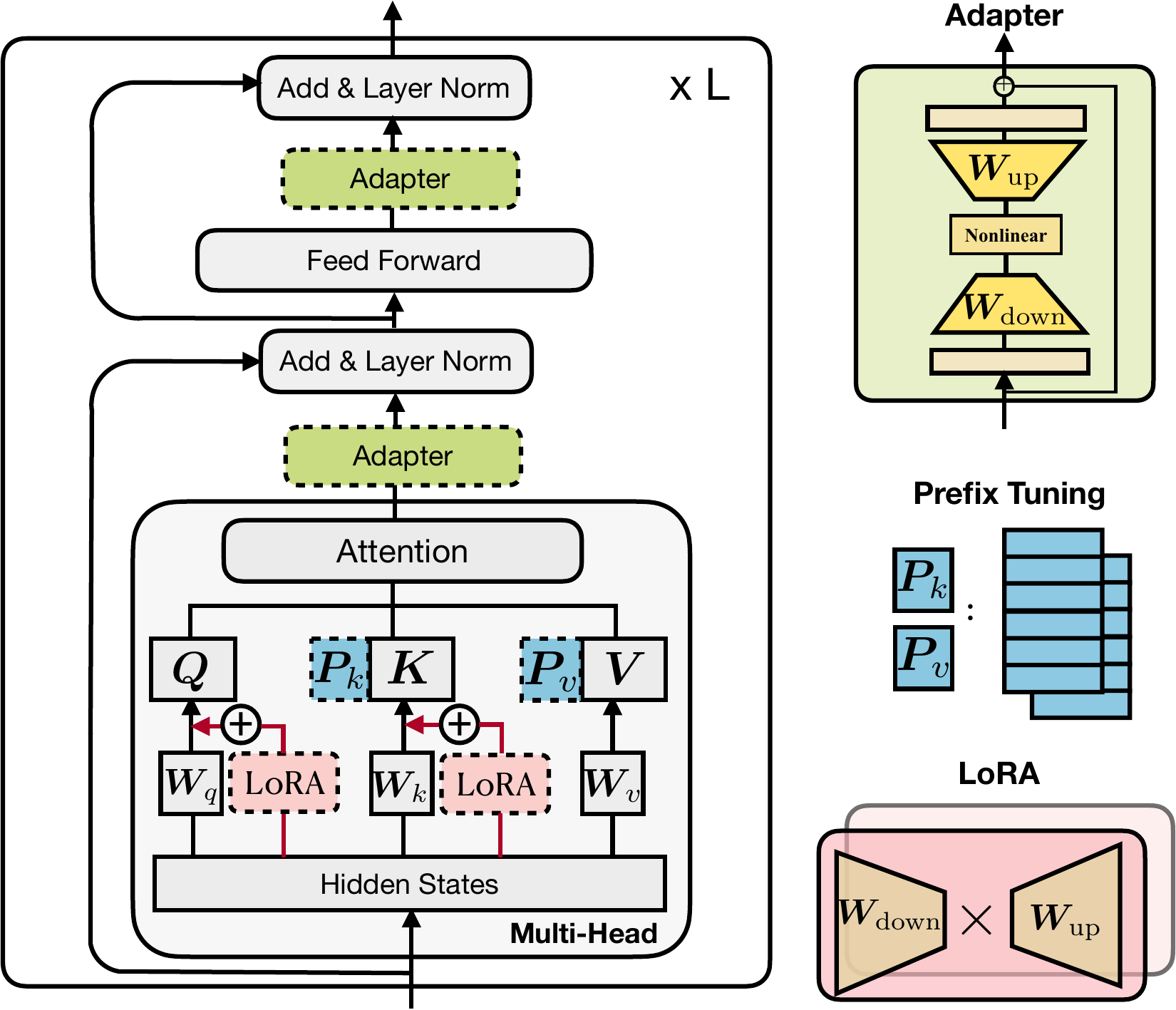}
        \caption{Illustration of the transformer architecture and several state-of-the-art parameter-efficient tuning methods. We use blocks with dashed borderlines to represent the added modules by those methods. 
        % \gn{I found the color palette of this to be a little bit ugly with all of the yellow/brown colors. Maybe it would be good to either have all of the standard components of the transformer be gray (since everyone knows them already) and the modifications be different colors chosen according to some nice color palette (such as the ``Tol'' one at the bottom of the page here: \url{https://davidmathlogic.com/colorblind/}). You could also match the colors with Figure 2 then.} \gn{Also it felt a little bit strange that BitFit is the only one in Figure 2 that is not represented here.}
        }
        \label{fig:intro-model}
    \end{figure}
    \end{minipage}\hfill
    \begin{minipage}{0.46\textwidth}
    \begin{figure}[H]
        \centering
        \includegraphics[width=0.9\textwidth]{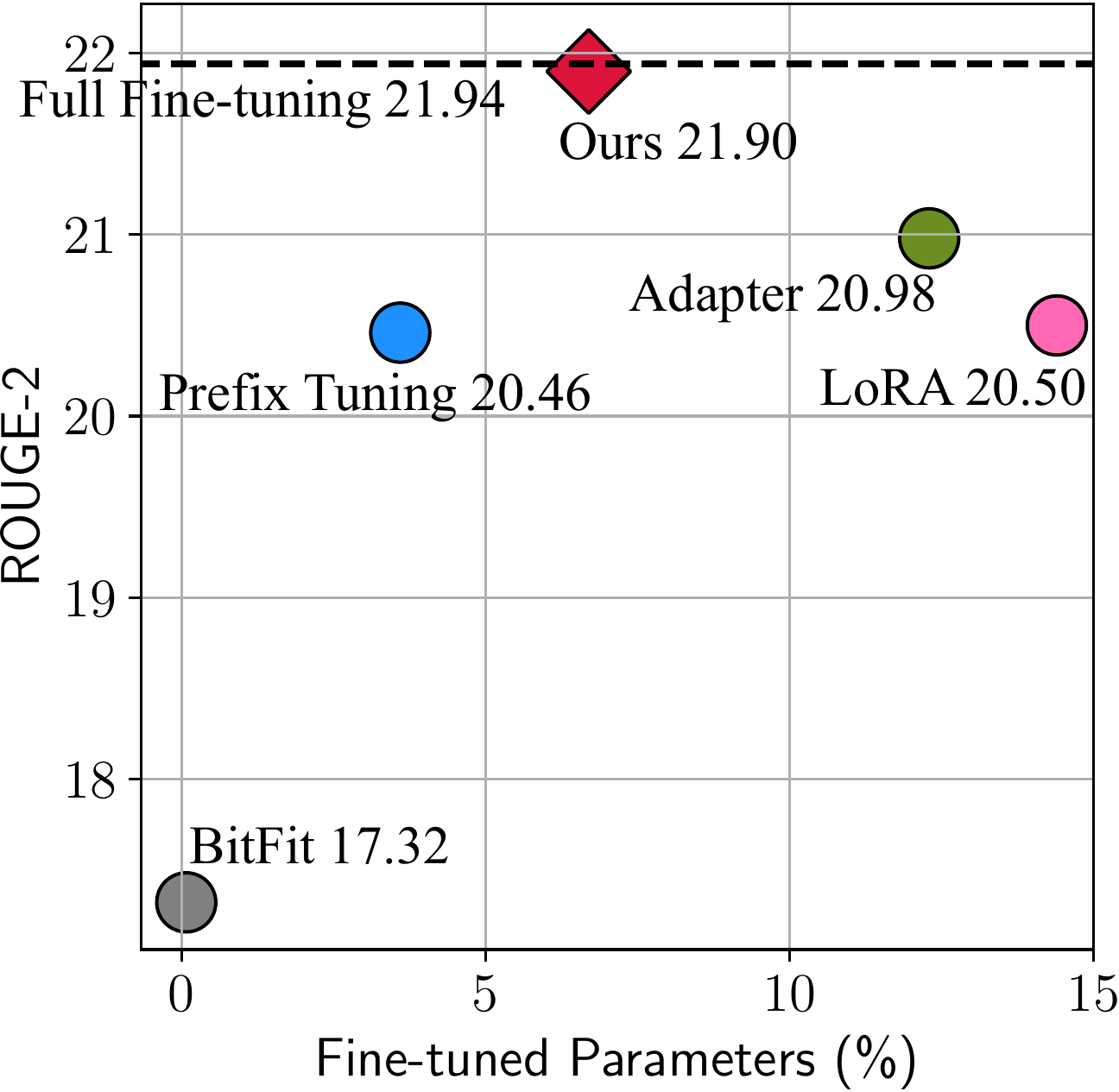}
        \caption{Performance of different methods on the XSum~\citep{xsum-emnlp} summarization task. The number of fine-tuned parameters is relative to the tuned parameters in full fine-tuning.
        }
        \label{fig:intro-res}
    \end{figure}
    \end{minipage}
        \vspace{-12pt}
\end{figure}

To mitigate this issue, a few lightweight alternatives have been proposed to update only a small number of extra parameters while keeping most pretrained parameters frozen. 
For example, \emph{adapter tuning}~\citep{houlsby2019parameter} inserts small neural modules called adapters to each layer of the pretrained network and only the adapters are trained at fine-tuning time. 
Inspired by the success of prompting methods that control PLMs through textual prompts~\citep{brown2020language,liu2021pre}, \emph{prefix tuning}~\citep{li2021prefix} and \emph{prompt tuning}~\citep{lester2021power} prepend an additional $l$ tunable prefix tokens to the input or hidden layers and only train these soft prompts when fine-tuning on downstream tasks. More recently,~\citet{hu2021lora} learn low-rank matrices to approximate parameter updates. We illustrate these methods in Figure~\ref{fig:intro-model}. These approaches have all been reported to demonstrate comparable performance to full fine-tuning on different sets of tasks, often through updating less than 1\% of the original model parameters. Besides parameter savings, parameter-efficient tuning makes it possible to quickly adapt to new tasks without catastrophic forgetting~\citep{pfeiffer2021adapterfusion} and often exhibits superior robustness in out-of-distribution evaluation~\citep{li2021prefix}. 

However, we contend that the important ingredients that contribute to the success of these parameter-efficient tuning methods are poorly understood, and the connections between them are still unclear. In this paper, we aim to answer three questions: (1) How are these methods connected? (2) Do these methods share design elements that are essential for their effectiveness, and what are they?
% (2) Do these methods perform differently in different tasks/settings? If so, what elements of their design make them superior in one setting and inferior in another? 
(3) Can the effective ingredients of each method be transferred to others to yield more effective variants? 
% (3) Can the effective ingredients of each method be combined to yield parameter-efficient tuning methods that are more effective overall? 

In order to answer these questions, we first derive an alternative form of prefix tuning that reveals prefix tuning's close connections with adapters (\textsection\ref{sec:connection}).
Based on this we then devise a unified framework that frames
% establishes connections among state-of-the-art parameter-efficient tuning methods. Specifically, 
% we focus on the self-attentional Transformer~\citep{vaswani2017attention} architecture adopted by most PLMs, 
% we unify 
the aforementioned methods as different ways to modify the hidden representations of frozen PLMs (\textsection\ref{sec:unify}).
Our unified framework decomposes previous methods along a \emph{shared} set of design dimensions, such as the function used to perform the modification, the position in which to impose this modification, and how to integrate the modification.
% \gn{``compose'' is not very clear here}.
This framework allows us to transfer design choices across approaches to propose new variants 
% \gn{specifically what do you propose? I don't think you need to describe in detail, but it'd be good to give at least an idea}
such as adapters with multiple heads (\textsection\ref{sec:transfer}). In experiments, we first show that existing parameter-efficient tuning methods still lag behind full fine-tuning on higher-resource and challenging tasks (\textsection\ref{sec:exp-previous-res}), as exemplified in Figure~\ref{fig:intro-res}. Then we utilize the unified framework to identify critical design choices and validate the proposed variants empirically (\textsection\ref{sec:exp:insertion}-\ref{sec:combine}). Our experiments on four NLP benchmarks covering text summarization, machine translation (MT), text classification, and general language understanding, 
% \gn{``language understanding'' and ``text classification'' are rather underspecified as tasks and there's actually a lot of overlap between them. Maybe we could write ``four NLP benchmarks covering text summarization, machine translation (MT), text classification, and general language understanding''}
demonstrate that the proposed variant uses less parameters than existing methods while being more effective, matching full fine-tuning results on all four tasks.

\section{Preliminaries}
% As background, we recap the equations of the transformer model, and review several parameter-efficient tuning.
\subsection{Recap of the transformer Architecture}
\label{sec:transformer}
% \begin{figure}[!t]
%     \centering
%     \includegraphics[width=0.5\textwidth]{figs/transformer.pdf}
%     \caption{Illustration of \jh{modify up proj as $\wu$ in adapter}}
%     \label{fig:overview}
%     \vspace{-10pt}
% \end{figure}

% Large-scale pretrained language models (PLMs) have demonstrated their remarkable capabilities in efficient transfer learning across various NLP tasks, such as classification~\citep{liu2019roberta}, machine translation~\citep{xue-etal-2021-mt5} and abstractive summarization~\citep{lewis-etal-2020-bart}. 
% Most PLMs, e,g. GPT-2~\citep{radford2019language}, BERT~\citep{devlin2019bert}, BART~\citep{lewis2020bart}, adopt the transformer~\citep{vaswani2017attention} architecture as shown in Fig.~\ref{fig:overview} (without the dashed boxes).
% \gn{I modified the first two sentences here to make it clear the motivation of this section, not that we're trying to teach people about the transformer, but because we need to define the equations here for later reference.}
The transformer model~\citep{vaswani2017attention} is now the workhorse architecture behind most state-of-the-art PLMs.
In this section we recap the equations of this model for completeness. 
% but we assume most readers are familiar with the basics already -- those who are not may refer to \citet{vaswani2017attention} for details.
Transformer models are composed of $L$ stacked blocks, where 
each block (Figure~\ref{fig:intro-model}) contains two types of sub-layers: multi-head self-attention and a fully connected feed-forward network (FFN).\footnote{In an encoder-decoder architecture, the transformer decoder usually has another multi-head cross-attention module between the self-attention and FFN, which we omit here for simplicity.} 
% We the number of queries as $n$ and the number of key-value pairs as $m$
% \gn{these concepts have not been mentioned yet, so this should be moved after the next sentence}.
The conventional attention function maps queries $\mQ\in \mathbb{R}^{n\times d_k}$ and key-value pairs $\mK \in \mathbb{R}^{m\times d_k}, \mV \in \mathbb{R}^{m\times d_v}$:
% where $n$ and $m$ are the number of queries and key-value pairs respectively.  
% \gn{I removed ``the'' from ``the queries'' and ``the key-value pairs'' to indicate that they haven't been mentioned yet.} to the output as:
\begin{equation}
    \mathrm{Attn}(\mQ, \mK, \mV) = \text{softmax}(\frac{\mQ\mK^T}{\sqrt{d_k}})\mV,
\end{equation}
where $n$ and $m$ are the number of queries and key-value pairs respectively.
Multi-head attention performs the attention function in parallel over $N_h$ heads, where each head is separately parameterized by $\mW_q^{(i)}, \mW_k^{(i)}, \mW_v^{(i)} \in \sR^{d\times d_h}$ to project inputs to queries, keys, and values. Given a sequence of $m$ vectors $\mC \in \sR^{m\times d}$ over which we would like to perform attention and a query vector $\vx\in\sR^d$, 
% single-head attention (SHA)~\citep{luong-etal-2015-effective} parameterized by $\wk, \wq, \wv, \wo \in \sR^{d\times d}$ module computes the output as:
% \begin{equation}
%     \text{SHA}(\mC, \vx)=\text{Attn}(\vx\wq, \mC\wk, \mC\wv)\wo .
% \end{equation}
multi-head attention (MHA) computes the output on each head and concatenates them:\footnote{Below, we sometimes ignore the head index $i$ to simplify notation when there is no confusion.}
\begin{equation}
\begin{aligned}
    \mathrm{MHA}(\mC, \vx) =  \mathrm{Concat(head_1, \cdots, head_h)}\wo, \
    \mathrm{head_i} = \mathrm{Attn}(\vx\mW_q^{(i)}, \mC\mW_k^{(i)}, \mC\mW_v^{(i)}), 
    \label{eq:multihead:attn}
\end{aligned}
\end{equation}
where $\wo \in \sR^{d\times d}$.
% \gn{I felt a little bit weird seeing $\mC$ here, although I can see why you did this. I wonder if there are better variable names?}
$d$ is the model dimension, and 
in MHA $d_h$ is typically set to $d / N_h$ to save parameters, which indicates that each attention head is operating on a lower-dimensional space. 
% mechanism performs the attention function $N_h$ times in parallel on $N_h$ heads. Each head has different linear projections to project the queries, keys, and values to $d_k, d_k, d_v$ dimensions as input to the head attention. Then the outputs of the heads are concatenated and projected again:
% In multi-head self attention, $\mQ=\mK=\mV=X\in \sR^{n\times d_{\text{model}}}$ which is the set of hidden representations.  We note that in transformers $d_k=d_v=d_{model}/N_h$ 
The other important sublayer is the fully connected feed-forward network (FFN) which consists of two linear transformations with a ReLU activation function in between:
\begin{equation}
    \mathrm{FFN}(\vx) = \mathrm{ReLU}(\vx\mW_1 +\vb_1)\mW_2 + \vb_2,
\end{equation}
where $\mW_1 \in \mathbb{R}^{d\times d_m}$, $\mW_2 \in \sR^{d_m\times d}$. Transformers typically use a large $d_m$, e.g. $d_m=4d$. 
Finally, a residual connection is used followed by layer normalization~\citep{ba2016layer}.

\subsection{Overview of Previous Parameter-efficient Tuning Methods}
\label{sec:overview}
Below and in Figure~\ref{fig:intro-model}, we introduce several state-of-the-art parameter-efficient tuning methods. 
Unless otherwise specified, they only tune the added parameters while the PLM's are frozen. 
\paragraph{Adapters~\citep{houlsby2019parameter}:} The adapter approach inserts small modules (adapters) between transformer layers. The adapter layer generally uses a down-projection with $\wdd \in\sR^{d\times r}$ to project the input $\vh$ to a lower-dimensional space specified by bottleneck dimension $r$, followed by a nonlinear activation function $f(\cdot)$, and a up-projection with $\wu\in\sR^{r\times d}$.
These adapters are surrounded by a residual connection, 
% \gn{add if space allows? ``, allowing the original model's information to be passed through as-is, with the adapter only making changes to the original representations''}
leading to a final form:
\begin{equation}
\label{eq:adapter}
    \vh \leftarrow \vh + f(\vh\wdd)\wu.
\end{equation}
\citet{houlsby2019parameter} places two adapters sequentially within one layer of the transformer, one after the multi-head attention and one after the FFN sub-layer.
% Adapters modify the hidden states $\vh$ of the inserted position as where $f$ is a parameter-free nonlinear function like ReLU.
\citet{pfeiffer2021adapterfusion} have proposed a more efficient adapter variant that is inserted only after the FFN ``add \& layer norm" sub-layer.

\paragraph{Prefix Tuning~\citep{li2021prefix}:}
% \gn{It seems a bit strange to list Prompt Tuning in parallel as Prefix Tuning given that we only compare with Prefix Tuning. Also, why is P-Tuning not included then? Maybe just make Prefix Tuning be the title of the paragraph, and mention Prompt Tuning and P-Tuning at the end?}
Inspired by the success of textual prompting methods \citep{liu2021pre}, 
% recent works have proposed to learn trainable, soft prompt tokens by backpropagating gradients from the task objective function. 
prefix tuning prepends $l$ tunable prefix vectors to the keys and values of the multi-head attention at every layer. Specifically, two sets of prefix vectors $\pk, \pv \in \sR^{l\times d}$ are concatenated with the original key $\mK$ and value $\mV$. Then multi-head attention is performed on the new prefixed keys and values. The computation of $\mathrm{head}_i$ in Eq.~\ref{eq:multihead:attn} becomes:
\begin{equation}
\label{eq:pt}
    \mathrm{head}_i = \mathrm{Attn}(\vx\mW_q^{(i)}, \text{concat}(\pk^{(i)}, \mC\wk^{(i)}), \text{concat}(\pv^{(i)}, \mC\wv^{(i)})), 
\end{equation}
$\pk$ and $\pv$ are split into $N_h$ head vectors respectively and $\pk^{(i)}, \pv^{(i)} \in \sR^{l\times d/N_h}$ denote the $i$-th head vector. Prompt-tuning~\citep{lester2021power} simplifies prefix-tuning by only prepending to the input word embeddings in the first layer; similar work also includes P-tuning~\citep{liu2021gpt}.

\paragraph{LoRA~\citep{hu2021lora}:} LoRA injects trainable low-rank matrices into transformer layers to approximate the weight updates. For a pre-trained weight matrix $\mW \in \sR^{d\times k}$, LoRA represents its update with a low-rank decomposition $\mW + \Delta W = \mW + \wdd\wu$, where $\wdd \in \sR^{d\times r}, \wu \in \sR^{r\times k}$ are tunable parameters.
% \footnote{With a bit abuse of notations, we use $r$, $\wu$, $\wdd$ in different methods. \gn{It wasn't clear what this footnote was trying to say? Are you trying to say that the notation is the same as the adapters but these are different methods? I think this is pretty obvious, so maybe this footnote could be removed to save space.}}
LoRA applies this update to the query and value projection matrices $(\wq, \wv)$ in the multi-head attention sub-layer, as shown in Figure~\ref{fig:intro-model}. For a specific input $\vx$ to the linear projection in multi-head attention, LoRA modifies the projection output $\vh$ as: 
\begin{equation}
\label{eq:lora}
    \vh \leftarrow \vh + s\cdot\vx\wdd\wu,
\end{equation}
where 
% $\vh$ is the projected query or value vector and 
$s \ge 1$ is a tunable scalar hyperparameter.\footnote{The public code of LoRA at \href{https://github.com/microsoft/LoRA}{https://github.com/microsoft/LoRA} uses different $s$ in different datasets, and we have verified the value of $s$ could have a significant effect on the results.}
% \footnote{While~\cite{hu2021lora} did not mention this tunable hyperparameter in the preprint at the time of this submission (10/05/2021), their latest code at \url{https://github.com/microsoft/LoRA} tunes this hyperparameter, and we have verified that this has a significant effect on results.}

\paragraph{Others:} Other parameter-efficient tuning methods include BitFit~\citep{ben2021bitfit}, which only fine-tunes bias vectors in the pre-trained model, and diff-pruning~\citep{guo2020parameter}, which learns a sparse parameter update vector.
\section{Bridging the Gap -- A Unified View}
We first derive an equivalent form of prefix tuning to establish its connection with adapters. We then propose a unified framework for parameter-efficient tuning that includes several state-of-the-art methods as instantiations.
% Our unified lens formulates different parameter-efficient tuning methods as combination of a set of shared design components, which help further understand the key ingredients to make them work well.

\begin{figure}[!t]
    \centering
    \begin{subfigure}[b]{0.19\textwidth}
    	\includegraphics[width=\textwidth]{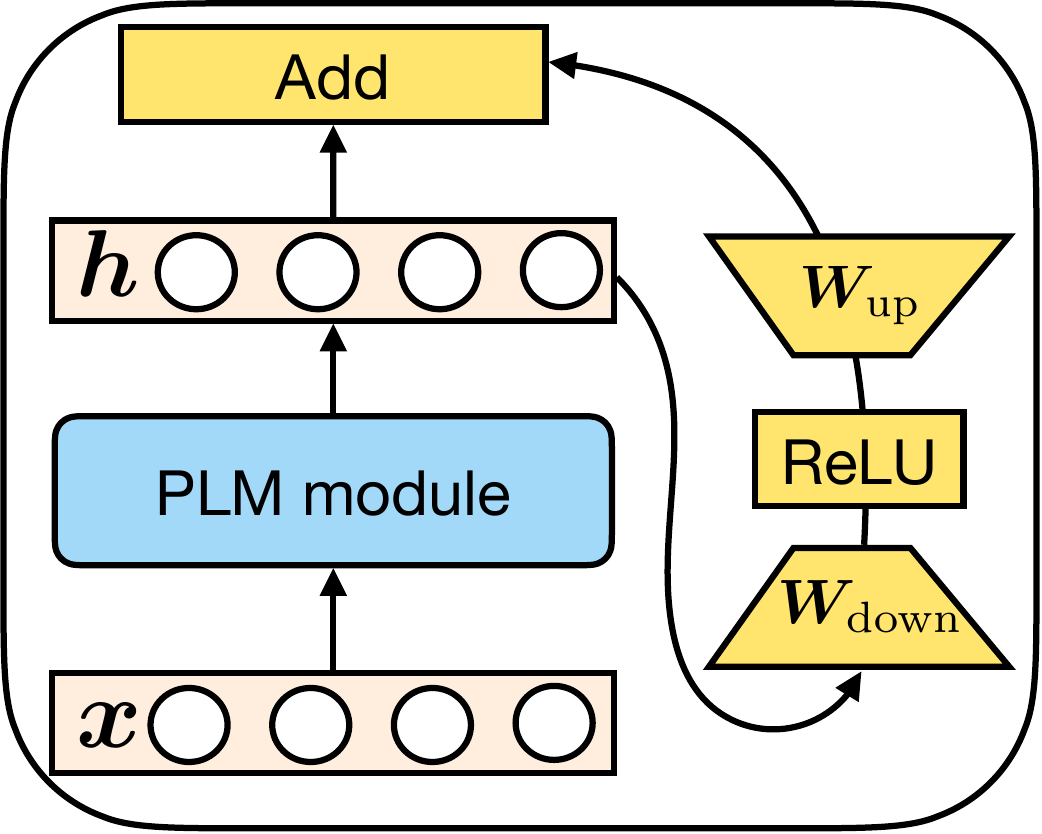}
    	\caption{Adapter}
    	\label{fig:adapter}
    \end{subfigure}
    \hfill
    \begin{subfigure}[b]{0.19\textwidth}
    	\includegraphics[width=\textwidth]{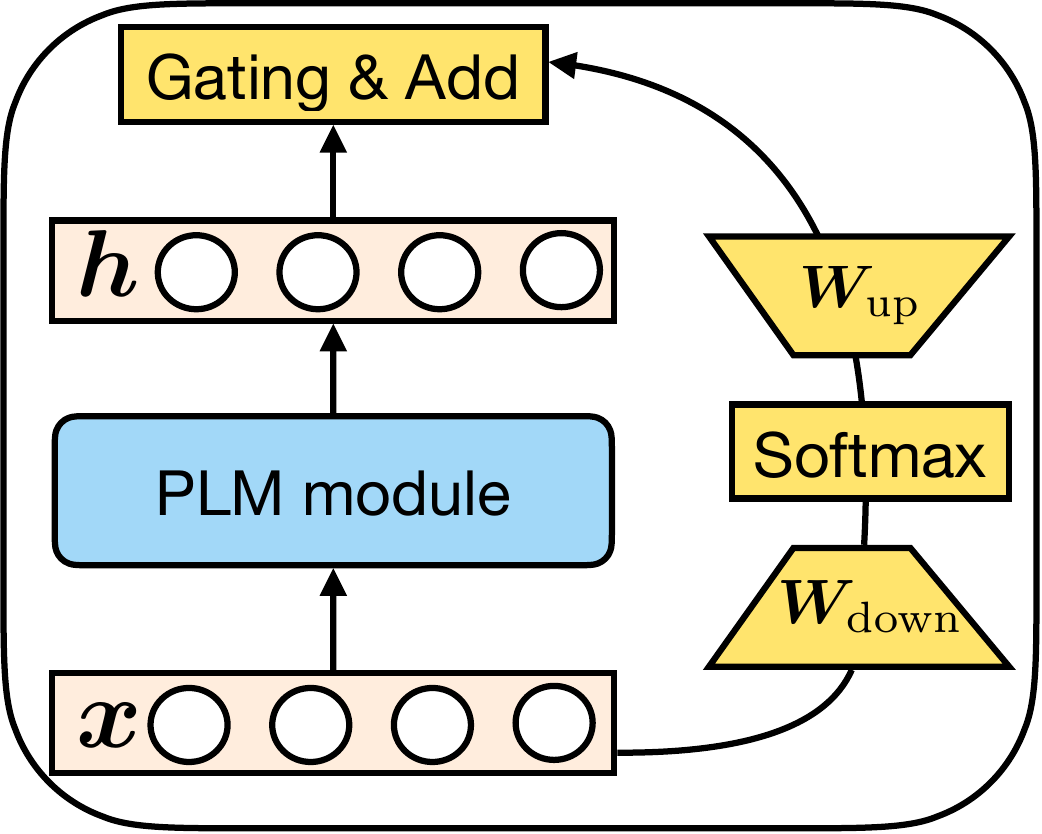}
    	\caption{Prefix Tuning}
    	\label{fig:prefix1}
    \end{subfigure}
    \hfill
    \begin{subfigure}[b]{0.19\textwidth}
    	\includegraphics[width=\textwidth]{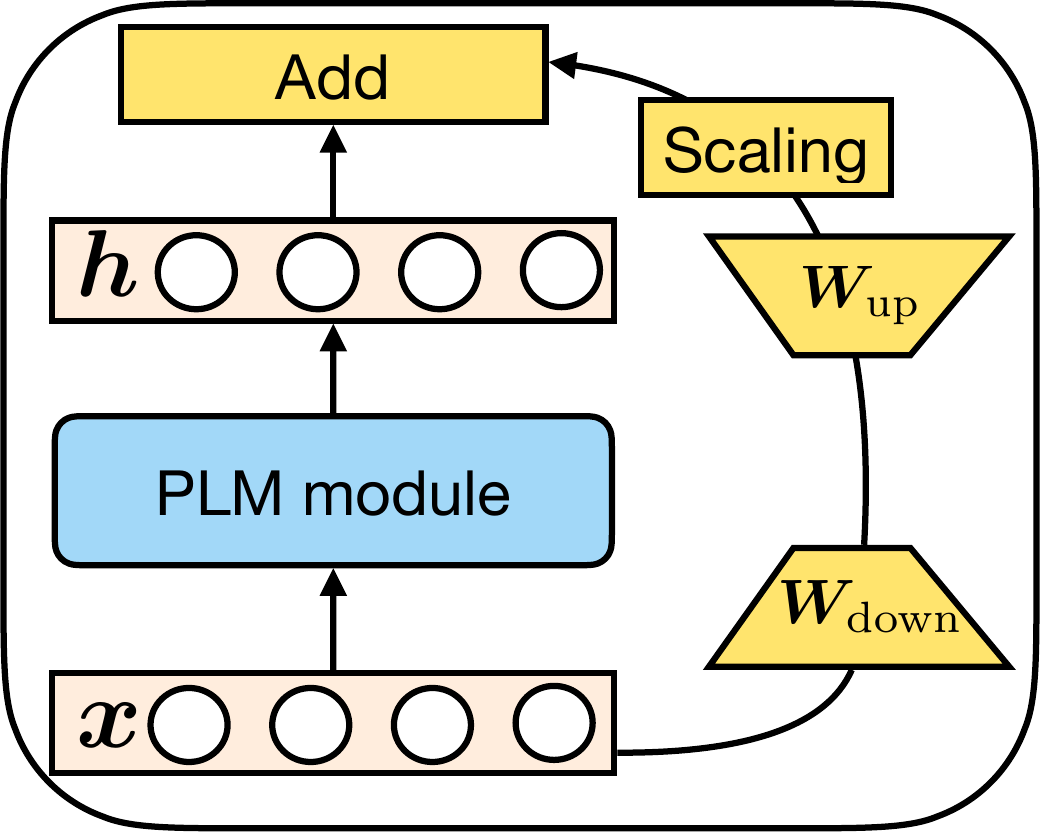}
    	\caption{LoRA}
    	\label{fig:lora}
    \end{subfigure}
    \hfill
     \begin{subfigure}[b]{0.19\textwidth}
    	\includegraphics[width=\textwidth]{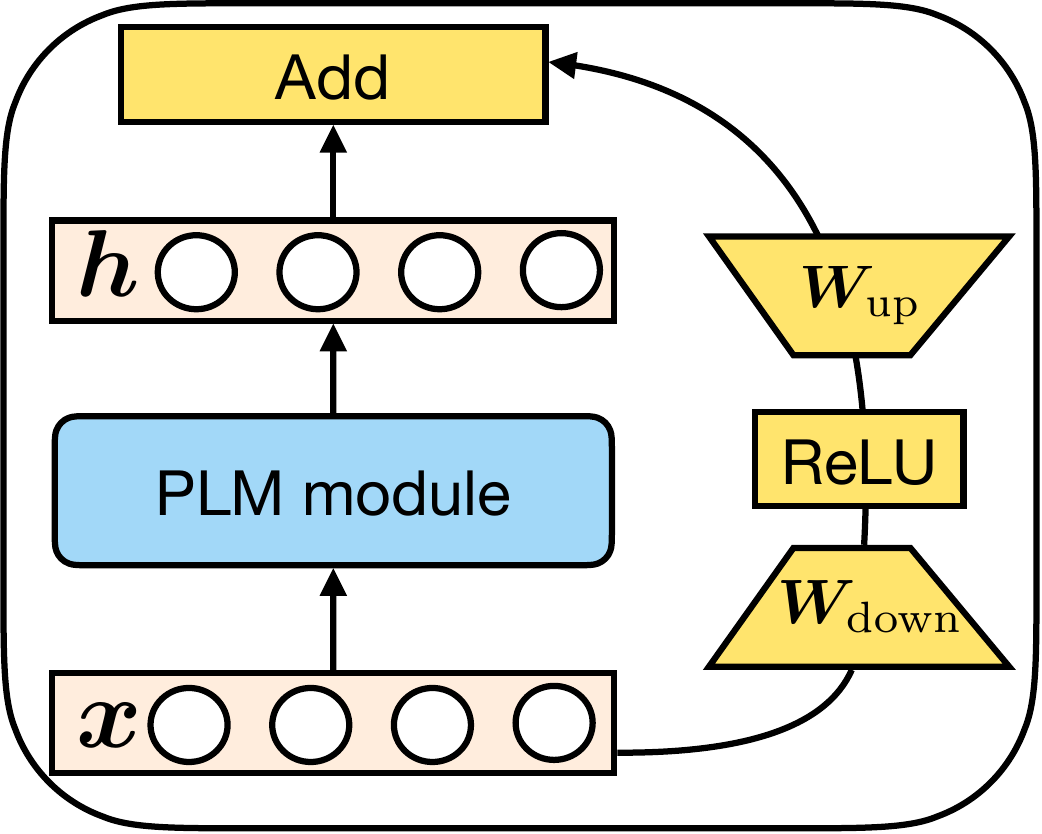}
    	\caption{Parallel Adapter}
    	\label{fig:parallel}
    \end{subfigure}
    \hfill
     \begin{subfigure}[b]{0.19\textwidth}
    	\includegraphics[width=\textwidth]{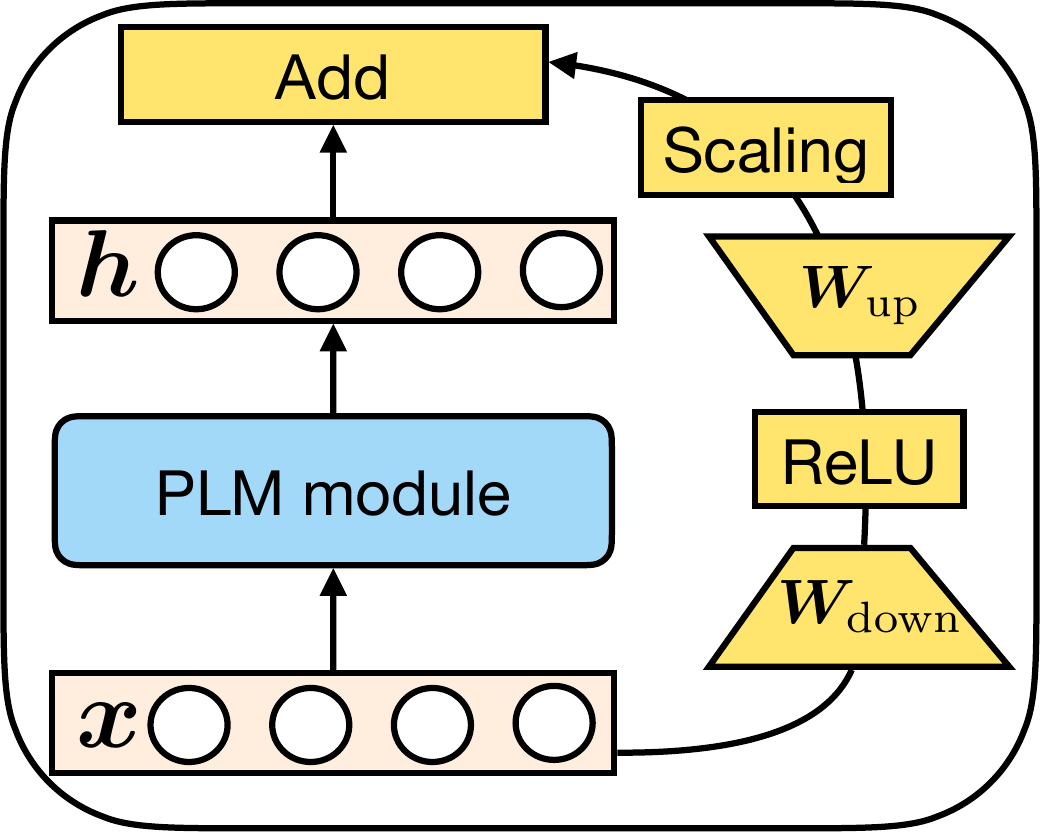}
    	\caption{Scaled PA}
    	\label{fig:scale}
    \end{subfigure}
    \vspace{-5pt}
    \caption{Graphical illustration of existing methods and the proposed variants. ``PLM module'' represents a certain sublayer of the PLM (e.g. attention or FFN) that is frozen.
    % add me back
    ``Scaled PA'' denotes scaled parallel adapter. We do not include multi-head parallel adapter here to save space.
    }
    \vspace{-12pt}
    \label{fig:connection}
\end{figure}

\subsection{A Closer Look at Prefix Tuning}
\label{sec:connection}
\label{sec:reinter-prefix}
Eq.~\ref{eq:pt} describes the mechanism of prefix tuning which changes the attention module through prepending $l$ learnable vectors to the original attention keys and values. Here, we derive an equivalent form of Eq.~\ref{eq:pt} and provide an alternative view of prefix tuning:\footnote{Without loss of generalization, we ignore the softmax scaling factor $\sqrt{d}$ for ease of notation.}
\begin{equation}
\label{eq:prefix-adapter}
\begin{split}
& \text{head} = \text{Attn}(\vx\wq, \text{concat}(\pk, \mC\wk), \text{concat}(\pv, \mC\wv)) \\
& = \text{softmax}\big(\vx\wq\tc(\pk, \mC\wk)^\top\big) \begin{bmatrix} \pv \\ \mC\wv \end{bmatrix} \\
& = (1 - \lambda(\vx)) \text{softmax}(\vx\wq\wk^\top\mC^\top)\mC\wv + \lambda(\vx)\text{softmax}(x\wq\pk^\top)\pv \\
& = (1 - \lambda(\vx)) \underbrace{ \text{Attn}(\vx\wq, \mC\wk, \mC\wv) }_{\text{standard attention}} + \lambda(\vx) \underbrace{ \text{Attn}(\vx\wq, \pk, \pv) }_{\text{independent of }\mC},
\end{split}
\end{equation}
where $\lambda(\vx)$ is a scalar that represents the sum of normalized attention weights on the prefixes:
\begin{equation}
\lambda(\vx) = \frac{\sum_i\exp (\vx\wq\pk^\top)_i}{\sum_i \exp (\vx\wq\pk^\top)_i + \sum_j \exp(\vx\wq\wk^\top\mC^\top)_j}.
\end{equation} 
Note that the first term in Eq.~\ref{eq:prefix-adapter}, $\text{Attn}(\vx\wq, \mC\wk, \mC\wv)$, is the original attention without prefixes, whereas the second term is a position-wise modification independent of $\mC$. 
% Intuitively, it indicates that prefix tuning actually down-weights the original attention probabilities by a scalar factor (i.e. $1-\lambda$), and redistribute the remaining attention probability mass $\lambda$ to attend to prefixes.
Eq.~\ref{eq:prefix-adapter} gives an alternative view of prefix tuning that essentially applies a position-wise modification to the original head attention output $\vh$ through linear interpolation:
\begin{equation}
\label{eq:prefix-final}
\vh \leftarrow (1-\lambda(\vx))\vh + \lambda(\vx)\Delta\vh, \quad \Delta\vh := \text{softmax}(\vx\wq\pk^\top)\pv.
\end{equation}
% \vspace{-15pt}
\paragraph{The Connection with Adapters:}
We define $\mW_1$=$\wq\pk^\top$, $\mW_2$=$\pv$, $f$=$\text{softmax}$, and rewrite Eq.~\ref{eq:prefix-final}:
\begin{equation}
\label{eq:prefix-final2}
\vh \leftarrow (1-\lambda(\vx))\vh + \lambda(\vx) f(\vx\mW_1)\mW_2,
\end{equation}
which reaches a very similar form to the adapter function in Eq.~\ref{eq:adapter}, except that prefix tuning is performing weighted addition while the adapter one is unweighted.\footnote{
% Adapters and prefix tuning are generally applied to different positions, and thus $\vh$ can differ,
$\vh$ in adapters and prefix tuning are usually different,
as described more below. However, here we mainly discuss the functional form as adapters can, in principle, be inserted at any position.
} Figure~\ref{fig:prefix1} demonstrates the computation graph of prefix tuning from this view, which allows for abstraction of prefix tuning as a plug-in module like adapters. Further, we note that $\mW_1 \in \sR^{d_h\times l}$ and $\mW_2 \in \sR^{l\times d_h}$ are low-rank matrices when $l$ is small, and thus they function similarly to the $\wdd$ and $\wu$ matrices in adapters. 
% To compute the modification vector $\Delta \vh$, the input is first projected to low-dimensional space through $\mW_1$, then passed a nonlinear function $f$ (softmax in this case), and projected back to $d$-dimensional vector through $\mW_2$ -- such a mechanism is exactly the same as adapters. 
This view also suggests that the number of prefix vectors, $l$, plays a similar role to the bottleneck dimension $r$ in adapters: they both represent the rank limitation of computing the modification vector $\Delta \vh$. Thus we also refer $l$ as the bottleneck dimension. Intuitively, the rank limitation implies that $\Delta \vh$ is a linear combination of \emph{the same} $l$ (or $\le l$) basis vectors
% (e.g. row vectors of $\pv$) 
for any $\vx$.  

\paragraph{The Difference from Adapters:}
In addition to the gating variable $\lambda$, we emphasize three differences between prefix tuning and adapters.
(1) As demonstrated in Figure~\ref{fig:connection}, prefix tuning uses $\vx$, the input of the PLM layer, to compute $\dvh$, while adapters use $\vh$, the output of the PLM layer.
Thus, prefix tuning can be thought of as a ``parallel'' computation to the PLM layer, whereas the typical adapter is ``sequential'' computation.
(2) Adapters are more flexible with respect to where they are inserted than prefix tuning: adapters typically modify attention or FFN outputs, while prefix tuning only modifies the attention output of each head.
Empirically, this makes a large difference as we will show in \textsection\ref{sec:exp-position}.
(3)
% \gn{I didn't fully understand this part. Could you try to re-read it and be a bit more concrete?}
% \jh{Need to make sure this is clear, comments welcome}
Eq.~\ref{eq:prefix-final2} applies to each attention head, while adapters are always single-headed, which makes prefix tuning more expressive: 
head attention is of dimension $d/N_h$ -- basically we have full rank updates to each attention head if $l \ge d / N_h$, but we only get full-rank updates to the whole attention output with adapters if $r \ge d$.
% because attention head is of dimension $d / N_h$, prefix tuning only needs $l\ge d / N_h$ to allow to compute flexible modification vectors while adapters needs $r \ge d$.  
% assume we have $n$ query vectors and the respective modification vectors form a matrix $\Delta \mH$, which is of dimension $n \times d / N_h$ in prefix tuning and $n \times d$ in adapters. The rank of $\Delta \mH$ is at most $r$ (or $l$). Prefix tuning allows full-rank modification $\Delta \mH$ to $\text{head}_i$ when $l\ge d / N_h$, and as a result $\text{concat}(\text{head}_i, \cdots, \text{head}_n)$ can be modified potentially into any value since each head and its prefixes are parameterized independently.  
% \footnote{The adapter function adds the same number of parameters as prefix tuning when $r=l$, as we will show in \textsection\ref{sec:exp-setup}.} 
% $\dvh$ is of dimension $d / N_h$ (as described in \textsection\ref{sec:transformer}) in prefix tuning yet $d$ in adapters, which implies that 
% prefix tuning can allow full-rank modification $\dvh$ for the head attention and as a result to the concatenated final attention when $l\ge d_h$, adapters, however, need $r \ge d$ to avoid rank limitation which requires more parameters than prefix tuning. 
% Adapters, however, need $r \ge d$ to achieve this. 
Notably, prefix tuning is not adding more parameters than adapters when $l=r$.\footnote{We will detail in \textsection\ref{sec:exp-setup} the number of parameters added of different methods.} 
We empirically validate such multi-head influence in \textsection\ref{sec:exp-position}.

% Our re-interpretation of prefix tuning reveals its previously unknown connections with adapters, and more importantly, leads to a unified framework for parameter-efficient tuning methods which we elaborate on next. \jh{maybe mention that many work that learns prompts is like learning a special version of adapter weights}
% Interestingly, while we derive the parallel adapter by drawing inspirations from prefix tuning, parallel adapter has been separately studied empirically in concurrent work for multilingual machine translation~\citep{}.

\vspace{-1mm}
\subsection{The Unified Framework}
\label{sec:unify}
\begin{table}[!t]
    \centering
    \caption{Parameter-efficient tuning methods decomposed along the defined design dimensions. Here, for clarity, we directly write the adapter nonlinear function as ReLU which is commonly used. 
    The bottom part of the table exemplifies new variants by transferring design choices of existing approaches.
    }
    \vspace{-5pt}
    \label{tab:design}
    \resizebox{1 \columnwidth}{!}{
    \begin{tabular}{lccccc}
    \toprule
     Method & $\dvh$ functional form & insertion form & modified representation & composition function\\
    \midrule
    \multicolumn{5}{c}{\bf Existing Methods} \\
    Prefix Tuning &  $\text{softmax}(\vx \wq \pk^{\top})\pv$ & parallel & head attn &$\vh \leftarrow (1-\lambda)\vh + \lambda\dvh$\\
    Adapter & $\text{ReLU}(\vh\wdd)\wu$ & sequential & ffn/attn & $\vh \leftarrow\vh + \dvh$\\
    LoRA & $\vx\wdd\wu$ & parallel & attn key/val &$\vh \leftarrow \vh + s\cdot\dvh$ \\
    \midrule
    \multicolumn{5}{c}{\bf Proposed Variants} \\
    Parallel adapter & $\text{ReLU}(\vh\wdd)\wu$ & parallel & ffn/attn & $\vh \leftarrow\vh + \dvh$\\
    Muti-head parallel adapter & $\text{ReLU}(\vh\wdd)\wu$ & parallel & head attn & $\vh \leftarrow\vh + \dvh$\\ 
    Scaled parallel adapter & $\text{ReLU}(\vh\wdd)\wu$ & parallel & ffn/attn & $\vh \leftarrow \vh + s\cdot \dvh$\\ 
    \bottomrule
    \end{tabular}}
    \vspace{-12pt}
\end{table}
Inspired by the connections between prefix tuning and adapters, we propose a general framework that aims to unify several state-of-the-art parameter-efficient tuning methods. Specifically, we cast them as learning a modification vector $\dvh$, which is applied to various hidden representations. Formally, we denote the hidden representation to be directly modified as $\vh$, and the direct input to the PLM sub-module that computes $\vh$ as $\vx$ (e.g.~$\vh$ and $\vx$ can be the attention output and input respectively). To characterize this modification process, we define a set of design dimensions, and different methods can be instantiated by varying values along these dimensions. We detail the design dimensions below, and illustrate how adapters, prefix tuning, and LoRA fall along them in Table~\ref{tab:design}:
% \gn{Based on our previous discussion, I tried to come up with some more elegant terminology here. Please tell me what you think. I still haven't thought of a better name for ``Insertion Form'' but I'd like to.}:

% \vspace{1mm} 
\noindent \textbf{Functional Form} 
is the specific function that computes $\dvh$. We have detailed the functional form for adapters, prefix tuning, and LoRA in Eq.~\ref{eq:adapter},~\ref{eq:lora}, and ~\ref{eq:prefix-final2} respectively. The functional forms of all these methods are similar with a \texttt{proj\_down $\rightarrow$ nonlinear $\rightarrow$ proj\_up} architecture, while ``nonlinear'' degenerates to the identity function in LoRA. 

% \vspace{1mm} 
\noindent \textbf{Modified Representation}
indicates which hidden representation is directly modified.\footnote{Strictly speaking, all the hidden representations would be indirectly influenced by modifying the ones before them. Here we refer to the position being \emph{directly} modified by the added module.}
% As shown in Table~\ref{tab:design}, prefix-tuning modify the attention output of each head while adapters modify the aggregated attention output or the FFN output, and LoRA modifies the key and value vectors in attention module. 

% \vspace{1mm} 
\noindent \textbf{Insertion Form}
is how the added module is inserted into the network. As mentioned in the previous section and shown in Figure~\ref{fig:connection}, traditionally adapters are inserted at a position in a sequential manner, where both the input and output are $\vh$. Prefix tuning and LoRA -- although not originally described in this way -- turn out to be equivalent to a parallel insertion where $\vx$ is the input. 
% Such a insertion form actually has practical impact which we study in \textsection\ref{sec:exp:insertion}.

% \vspace{1mm} 
\noindent \textbf{Composition Function}
is how the modified vector $\dvh$ is composed with the original hidden representation $\vh$ to form the new hidden representation. For example, adapters perform simple additive composition, prefix tuning uses a gated additive composition as shown in Eq.~\ref{eq:prefix-final2}, and LoRA scales $\dvh$ by a constant factor and adds it to the original hidden representation as in Eq.~\ref{eq:lora}. 

We note that many other methods not present in Table~\ref{tab:design} fit into this framework as well. For example, prompt tuning modifies the head attention in the first layer in a way similar to prefix tuning, and various adapter variants~\citep{pfeiffer2021adapterfusion,mahabadi2021compacter} can be represented in a similar way as adapters. Critically, the unified framework allows us to study parameter-efficient tuning methods along these design dimensions, identify the critical design choices, and potentially transfer design elements across approaches, as in the following section. 
\vspace{-1mm}
\subsection{Transferring Design Elements}
\label{sec:transfer}
Here, and in Figure~\ref{fig:connection}, we describe just a few novel methods that can be derived through our unified view above by transferring design elements across methods: 
(1) \emph{Parallel Adapter}
is the variant by transferring the parallel insertion of prefix tuning into adapters. Interestingly, while we motivate the parallel adapter due to its similarity to prefix tuning, concurrent work~\citep{zhu2021serial} independently proposed this variant and studied it empirically; (2) \emph{Multi-head Parallel Adapter} is a further step to make adapters more similar to prefix tuning: we apply parallel adapters to modify head attention outputs as prefix tuning. 
% \gn{This sentence is not very clear, it sounds like you're saying that this method constrains the position where the adapter is applied, when actually the difference is that it's multi-head, right? Also, why do you need to constrain the position? It seems this could be applied anywhere, right?} \cz{this actually couldn't be applied anywhere, because to use multi-head attention, we need to have the query/key/value projection matrix, and the original attention layer naturally has them. But at FFN, this needs to be learned, and the results are not very good. maybe we could make it more clear}. 
This way the variant improves the capacity for free by utilizing the multi-head projections as we discuss in \textsection\ref{sec:connection}. 
% We highlight that the only differences from prefix tuning are the gating composition and the nonlinear function;
%
(3) \emph{Scaled Parallel Adapter} is the variant by transferring the composition and insertion form of LoRA into adapters, as shown in Figure~\ref{fig:scale}.

Our discussion and formulation so far raise a few questions: Do methods varying the design elements above exhibit distinct properties? Which design dimensions are particularly important? Do the novel methods described above yield better performance? We answer these questions next. 
% \subsection{The Bottleneck Dimension}

%!TEX root=./iclr2022_conference.tex
\section{Experiments}
\label{sec:exp}

% Our experiments below are designed to (1) identify critical design choices through studying parameter-efficient methods along the design dimensions, and (2) evaluate new method variants. 

\vspace{-1mm}
\subsection{General Setup}
\label{sec:exp-setup}
\paragraph{Datasets:}
We study four downstream tasks: (1) XSum~\citep{xsum-emnlp} is an English summarization dataset
where models predict a summary given a news article; 
% There are 204K training samples; 
(2) English to Romanian translation using the WMT 2016 en-ro  dataset~\citep{bojar2016findings};
% contains 610K translations between English and Romanian, we focus on translating English into Romanian; 
(3) MNLI~\citep{N18-1101} is an English natural language inference dataset where models predict whether one sentence entails, contradicts, or is neutral to another.
% There are 393K training examples; 
(4) SST2~\citep{socher2013recursive} is an English sentiment classification benchmark where models predict whether a sentence's sentiment is positive or negative. 
% There are 67K training samples. 
% We obtain the MNLI and SST2 datasets from the GLUE benchmark~\citep{wang2018glue}.

\paragraph{Setup:}
We use \texttt{BART$_\text{LARGE}$}~\citep{lewis-etal-2020-bart} and a multilingual version of it, \texttt{mBART$_\text{LARGE}$}~\citep{TACL2107}, as the underlying pretrained models for XSum and en-ro translation respectively, and we use \texttt{RoBERTa$_\text{BASE}$}~\citep{liu2019roberta} for MNLI and SST2. 
% The pretrained model parameters are frozen during parameter-efficient tuning except for bitfit where it fine-tunes the bias parameters of the pretrained models.
We vary the bottleneck dimension within $\{1, 30, 200, 512, 1024\}$ if needed.\footnote{
% % We choose 200 as the median value since it achieves the best results for prefix tuning on XSum as shown in~\citep{li2021prefix}. 
In some settings we use other values to match the number of added parameters of different methods.}
% we test bottleneck dimension 1024 only when the modified representation is FFN, because the training of prefix tuning does not fit into 48GB GPU memory when $l=1024$.\footnote{While other methods do not have memory issues, we keep the bottleneck dimension of attention modification at most 512 to have a relatively fair comparison with prefix tuning.} 
We mainly study adapters, prefix tuning (prefix), and LoRA which greatly outperform bitfit and prompt tuning in our experiments. 
In the analysis sections (\S\ref{sec:exp:insertion}-\ref{sec:exp-comp}) we insert adapters \emph{either} at the attention or FFN layers for easier analysis, but include the results of inserting at both places in the final comparison (\S\ref{sec:combine}).  
We re-implement these methods based on their respective public code.\footnote{
% We refer the respective official public code of adapter, prefix tuning, and LoRA to re-implement them, and we implement bitfit and prompt tuning on our own. 
We verify that our re-implementation can reproduce adapter and prefix tuning on XSum, and LoRA on MNLI, by comparing with the results of running the original released code.} We use the huggingface transformers library~\citep{wolf-etal-2020-transformers} for our implementation. Complete setup details can be found in Appendix~\ref{app:sec:exps}. 
\paragraph{Evaluation:}
We report ROUGE 1/2/L scores (R-1/2/L, ~\citet{lin-2004-rouge}) on the XSum test set, BLEU scores~\citep{papineni-etal-2002-bleu} on the en-ro test set, and accuracy on the MNLI and SST2 dev set. For MNLI and SST2, we take the median of five random runs. We also report the number of tuned parameters relative to that in full fine-tuning (\#params). 
% \gn{This is probably only medium priority, but it'd be better to run significance tests. Bootstrapping is the easiest way to do this. If you send me your outputs and the commands that you ran over them to do evaluation I could probably set it up if you don't have time.}

\paragraph{Number of Tunable Parameters:}
% add me back
% We compute the number of tunable parameters based on where the tunable module is inserted into and how it is parameterized.
% The pretrained-models for XSum or MT 
BART and mBART have an encoder-decoder structure that has three types of attention: encoder self-attention, decoder self-attention, and decoder cross-attention. 
RoBERTa only has encoder self-attention.
For each attention sub-layer, the number of parameters used of each method is: (1) prefix tuning prepends $l$ vectors to the keys and values and uses $2 \times l \times d$ parameters; (2) adapter has $\wdd$ and $\wu$ thus uses $2 \times r \times d$ parameters; (3) LoRA employs a pair of $\wdd$ and $\wu$ for query and value projections, hence uses $4 \times r \times d$ parameters. For the adapter modification at ffn, it uses $2 \times r \times d$ parameters which is the same as adapter at attention. 
Therefore, for a specific value of $r$ or $l$, 
prefix tuning uses the same number of parameters as adapters, while LoRA uses more parameters. More details can be found in Appendix~\ref{app:params}.
% \gn{Is this sentence necessary given that prompt tuning is not evaluated?}
% \gn{One thing I wasn't sure of: how did you calculate the denominator when calculating the percentage? Did you use only the body of the transformer network or did you count the word embeddings as well?} \cz{prompt-tuning is compared at the end, and we count all the parameters including the word embeddings.}

\subsection{The Results of Existing Methods}
\label{sec:exp-previous-res}
\begin{figure}[!t]
\begin{minipage}{.6\textwidth}
    \begin{figure}[H]
        \centering
        \begin{subfigure}[b]{0.49\columnwidth}        	\includegraphics[width=\columnwidth]{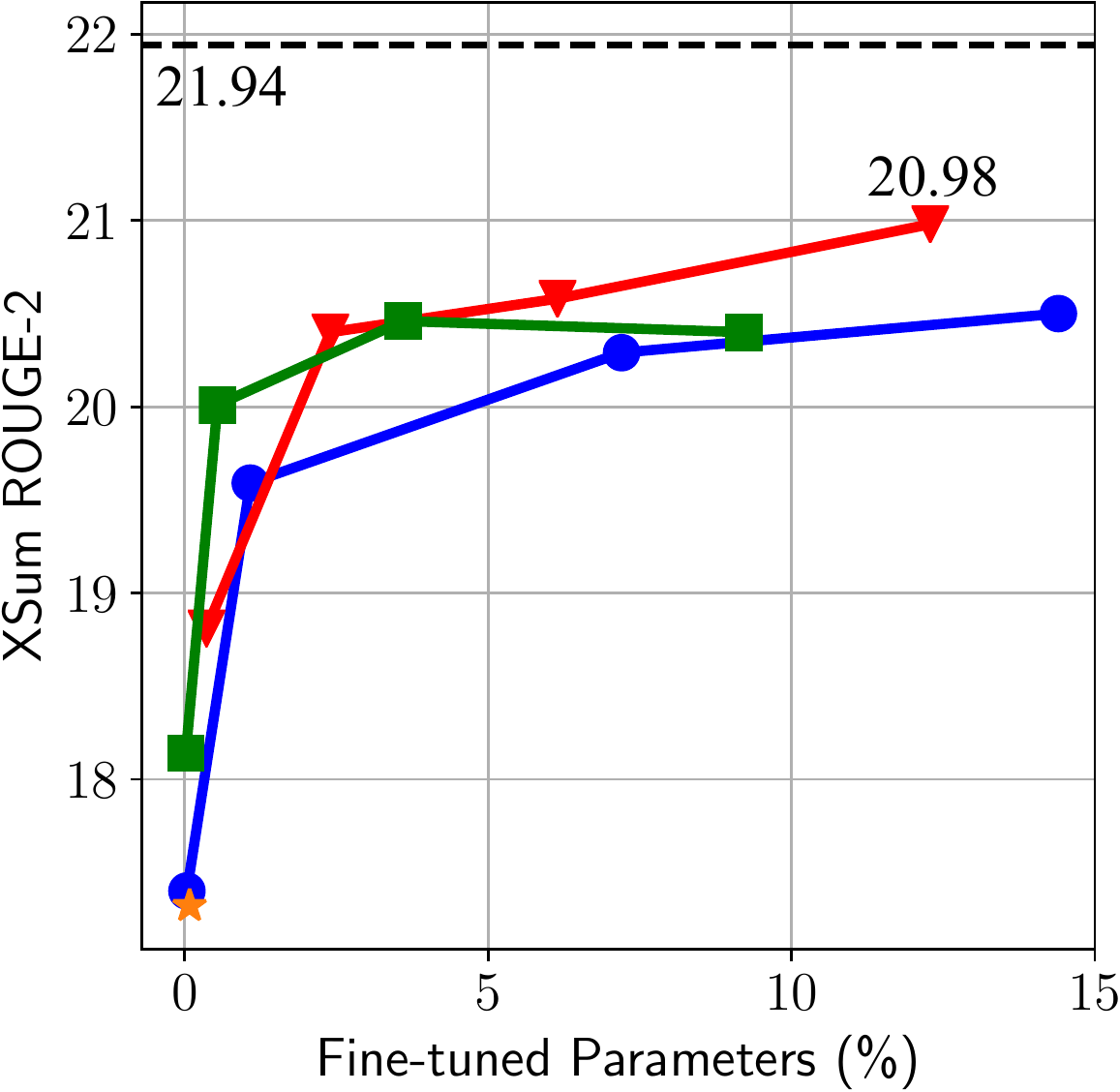}
        	\label{fig:res-xsum-overview}
        \end{subfigure}
        % \hfill
        \begin{subfigure}[b]{0.49\columnwidth}
    	\includegraphics[width=\columnwidth]{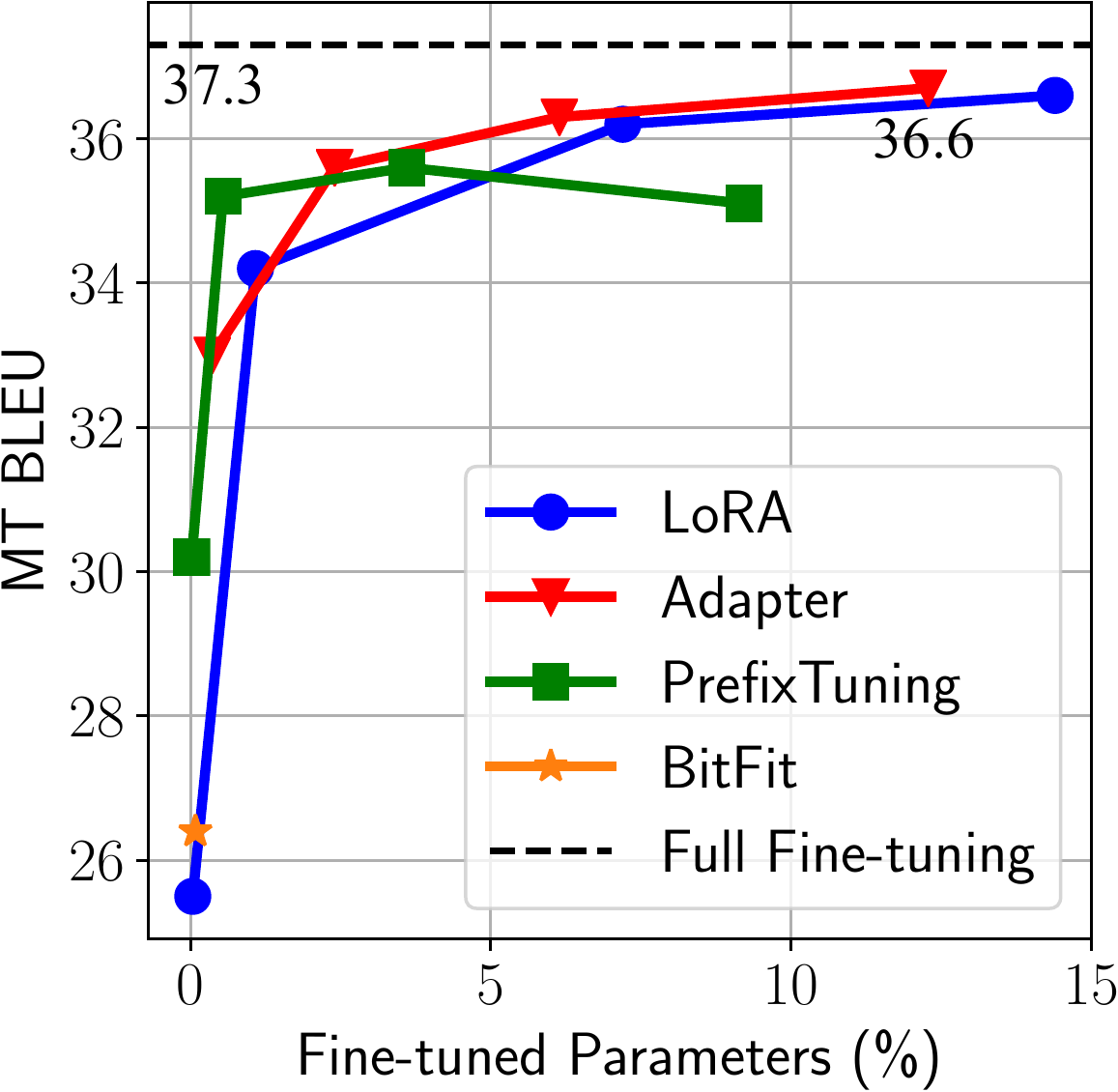}
    	\label{fig:res-mt-overview}
        \end{subfigure}
        \vspace{-5mm}
        \caption{Performance of previous state-of-the-art parameter-efficient tuning methods on XSum (left) and en-ro (right).\label{fig:res-overview2}}
    \end{figure}
\end{minipage}
\hfill
\begin{minipage}{.36\textwidth}
    \begin{table}[H]
    % \centering
    \caption{
    % Accuracy on the dev set MNLI and SST2. We report the overall (matched an mismatched) accuracy for MNLI.
    Accuracy on the dev set of MNLI and SST2.
    MAM Adapter is proposed in \textsection\ref{sec:combine}. Bitfit numbers are from~\citet{ben2021bitfit}. 
    % Results are the median from five random runs. 
    % These are all existing methods except MAM Adapter that is the proposed variant in \textsection\ref{sec:combine}
    }
    \label{tab:res-classification-overview}
    \vspace{-5pt}
    \resizebox{0.9 \columnwidth}{!}{
    \setlength{\tabcolsep}{2pt}
    \begin{tabular}{lcc}
    \toprule
     Method (\# params) & MNLI & SST2 \\
     \midrule
     Full-FT (100\%) & 87.6$_{\pm\text{.4}}$ & 94.6$_{\pm\text{.4}}$ \\
     \midrule
     Bitfit (0.1 \%) & 84.7 & 93.7 \\
     Prefix (0.5\%) & 86.3$_{\pm\text{.4}}$ & 94.0$_{\pm\text{.1}}$ \\
     LoRA (0.5\%)  & 87.2$_{\pm\text{.4}}$ & 94.2$_{\pm\text{.2}}$\\
     Adapter (0.5\%) & 87.2$_{\pm\text{.2}}$ & 94.2$_{\pm\text{.1}}$ \\
    %  Adapter (0.9\%) & 87.3$_{\pm\text{.2}}$ & {\bf 94.8}$_{\pm\text{.3}}$ \\
     \midrule
     MAM Adapter (0.5\%) & {\bf 87.4}$_{\pm\text{.3}}$& 94.2$_{\pm\text{.3}}$  \\
    \bottomrule
    \end{tabular}}
\end{table}
\end{minipage}
\vspace{-12pt}
\end{figure}

We first overview the results of existing methods on the four tasks. 
% For XSum and en-ro we vary the bottleneck dimension, obtaining a series of results. 
% since previous methods exhibit a large gap towards full fine-tuning when using small bottleneck dimensions.
As shown in Figure~\ref{fig:res-overview2} and Table~\ref{tab:res-classification-overview}, while existing methods can achieve competitive
% (sometimes comparable to full fine-tuning) 
performance on MNLI and SST2 by tuning fewer than 1\% parameters, a large gap is still present if we add 5\% parameters in XSum and en-ro. The gap remains significant even though we increase the relative parameter size to $>$10\%.
% (1 R-2 points on XSum and 0.7 BLEU points on en-ro). 
Even larger gaps have been observed in~\citet{raffel2020exploring} on high-resource MT tasks. 
% This is notable as it indicates that many methods that have been demonstrated to achieve comparable accuracy to full fine-tuning on standard classification-based benchmarks such as GLUE or lower-resource, relatively simple tasks~\citep{guo2020parameter,ben2021bitfit,mahabadi2021compacter} may not generalize to higher-resource and more challenging tasks.
This shows that many methods that claimed comparable results to full fine-tuning on the GLUE benchmark with an encoder-only model~\citep{guo2020parameter,ben2021bitfit,mahabadi2021compacter}, or on relatively simple generation benchmarks such as E2E~\citep{novikova-etal-2017-e2e} with an encoder-decoder model~\citep{li2021prefix}, may not generalize well to other standard benchmarks. The influencing factors could be complicated including the number of training samples, task complexity, or model architecture.
% (e.g. encoder-only v.s. encoder-decoder).
We thus advocate for future research on this line to report results on more diverse benchmarks to exhibit a more complete picture of
their performance profile.
Below, our analysis will mainly focus on the XSum and en-ro datasets to better distinguish different design choices. We note that these two benchmarks are relatively high-resource performed with an encoder-decoder model (BART), while
% thus the conclusions in \textsection\ref{sec:exp:insertion}-\ref{sec:exp-comp} may be not applicable to other settings. 
we will discuss the results on MNLI and SST2 with an encoder-only model (RoBERTa) in \textsection\ref{sec:combine}.
\subsection{Which Insertion Form -- Sequential or Parallel?}
\label{sec:exp:insertion}
\begin{figure}[!t]
\begin{minipage}{.57\textwidth}
    \begin{table}[H]
        \centering
        \caption{Comparison of different insertion forms for adapters, i.e. sequential adapter (SA) and parallel adapter (PA). We include the results of prefix tuning as a reference point.}
        \vspace{-5pt}
        \label{tab:res-seq-par}
        \resizebox{0.95 \columnwidth}{!}{
        \setlength{\tabcolsep}{2pt}
        \begin{tabular}{lccr}
        \toprule
         Method & \text{\#} params & XSum (R-1/2/L) & MT (BLEU) \\
         \midrule
         Prefix, $l$=200 & 3.6\% & 43.40/20.46/35.51& 35.6\\
        %  LoRA, $r$=200 & 7.2 & 43.09/20.29/35.37& 36.2\\
         SA (attn), $r$=200 & 3.6\% & 42.01/19.30/34.40&  35.3 \\
        % attn-SA, $r$=512 & 9.2 & 43.29/20.40/35.37 & 35.1 \\
    %     Seq Adapter, r=512 & & &  \cz{xsum, no ln} \\
    %     Seq Adapter, r=1024 & &  & \cz{xsum, no ln} \\
    %     \hdashline
    %     Parallel Adapter, r=512 & & & \\
    %     Parallel Adapter, r=1024 & & & \cz{xsum, no ln} \\
        SA (ffn), $r$=200 & 2.4\% & 43.21/19.98/35.08 & 35.6 \\
        % ffn-SA, $r$=512 & 6.1 & 43.72/20.75/35.64 & 36.3 \\
        \midrule
        PA (attn), $r$=200 & 3.6\% & 43.58/20.31/35.34 & 35.6\\
        % attn-PA, $r$=512 & 2.4 & 43.99/20.83/35.77 & 36.2\\
        PA (ffn), $r$=200 & 2.4\% & {\bf 43.93/20.66/35.63} & {\bf 36.4} \\
        % ffn-PA, $r$=512 & 6.1 & 44.35/20.98/35.98 & 37.1 \\
        \bottomrule
        \end{tabular}}
        % \vspace{-5mm}
    \end{table}
\end{minipage}
\hfill
\begin{minipage}{.4\textwidth}
    \begin{table}[H]
        \centering
        \caption{Results on en-ro dataset.
        % when using 0.1\% parameters.
        \vspace{-5pt}
        \label{tab:res-mh}}
        \resizebox{0.95 \columnwidth}{!}{
        \setlength{\tabcolsep}{2pt}
        \begin{tabular}{lcr}
        \toprule
         Method & \# params & MT (BLEU) \\
         \midrule
    %     \multicolumn{3}{c}{attention} \\
        PA (attn), $r$=200 & 3.6\% & 35.6\\
        Prefix, $l$=200 & 3.6\% & 35.6 \\
        MH PA (attn), $r$=200 & 3.6\% & 35.8\\
        \midrule 
         Prefix, $l$=30 & 0.1\% & 35.2\\
          \ \ -gating, $l$=30 & 0.1\% &34.9 \\
         PA (ffn), $r$=30 & 0.1\%& 33.0\\
         PA (attn), $r$=30 & 0.1\%& 33.7\\
         MH PA (attn), $r$=30 & 0.1\% & {\bf 35.3}\\
    %     \midrule \\
    %     
    %     Adapter (Hi), r=200 & & \\
    %     MH Adapter, r=200 & & \\
        \bottomrule
        \end{tabular}}
        % \vspace{-5mm}
    \end{table}
    \end{minipage}
    \vspace{-12pt}
\end{figure}
We first study the insertion form design dimension, comparing the proposed parallel adapter (PA) variant to the conventional sequential adapter (SA) over both the attention (att) and FFN modification. We also include prefix tuning as a reference point. As shown in Table~\ref{tab:res-seq-par}, prefix tuning, which uses parallel insertion, outperforms attention sequential adapters. Further, the parallel adapter is able to beat sequential adapters in all cases,\footnote{More results with different $r$ can be found in Appendix~\ref{app:results}, which exhibits similar observations.} 
with PA (ffn) outperforming SA (ffn) by 1.7 R-2 points on XSum and 0.8 BLEU points on en-ro respectively. Given the superior results of parallel adapters over sequential adapters, we focus on parallel adapter results in following sections.

\subsection{Which Modified Representation -- Attention or FFN?}
\label{sec:exp-position}
% As we have described in \S\ref{sec:unify}, attention and FFN are two modification positions where the hidden representations are directly modified.
\paragraph{Setup:}
We now study the effect of modifying different representations. 
We mainly compare attention and FFN modification.
For easier analysis we categorize methods that modifies any hidden representations in the attention sub-layer (e.g. the head output, query, etc) as modifying the attention module. We compare parallel adapters at attention and FFN and prefix tuning. We also transfer the FFN modification to LoRA to have a LoRA (ffn) variant for a complete comparison. Specifically, we use LoRA to approximate the parameter updates for the FFN
weights $\mW_1\in \sR^{d\times d_m}$ and
$\mW_2\in \sR^{d_m\times d}$. In this case $\wu$ in LoRA for $\mW_1$ (similar for $\wdd$ of $\mW_2$) would have dimensions of $r \times d_m$, where $d_m=4d$ as described in \textsection\ref{sec:transformer}. Thus we typically use smaller $r$ for LoRA (ffn) than other methods to match their overall parameter size in later experiments.  

\begin{figure}[!t]
    \centering
    \begin{subfigure}[b]{0.45\textwidth}
    	\includegraphics[width=0.9\textwidth]{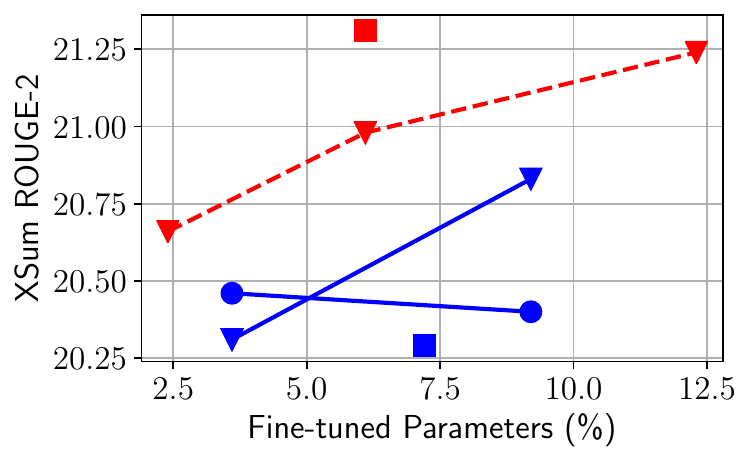}
    % 	\caption{}
    	\label{fig:xsum-position}
    \end{subfigure}
    % \hfill
    \hspace{10pt}
    \begin{subfigure}[b]{0.463\textwidth}
    	\includegraphics[width=0.9\textwidth]{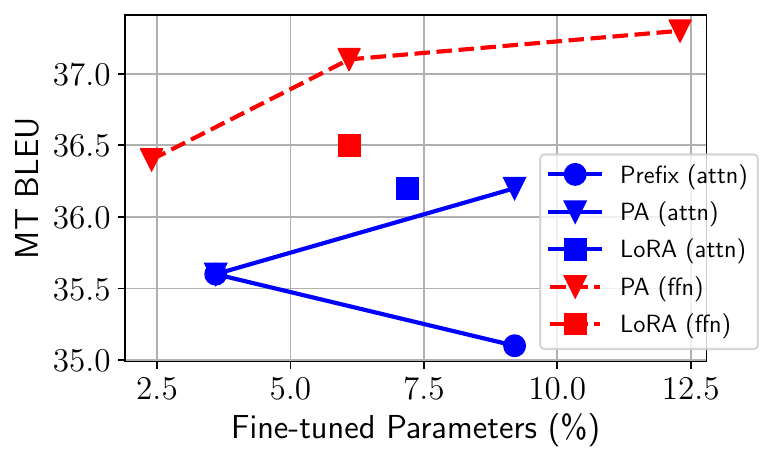}
    % 	\caption{}
    	\label{fig:mt-position}
    \end{subfigure}
    \vspace{-10pt}
    \caption{Results on XSum (left) and en-ro (right). PA represents parallel adapter. Blue and red markers 
    % \gn{you might consider making the text ``red'' and ``blue'' red and blue in the same shade as the figure, which also helps people who print papers in black and white, or are colorblind, tell which is which.} 
    apply modifications at attention and FFN sub-layers respectively (best viewed
    in color).
    }
    \label{fig:position}
    \vspace{-15pt}
\end{figure}

\paragraph{Results:}
% In Figure~\ref{fig:position}, we compare the results of applying the modification vectors to different positions. 
As shown in Figure~\ref{fig:position}, \emph{any} method with FFN modification outperforms \emph{all} the methods with attention modification in all cases (the red markers are generally above all the blue ones, the only exception is ffn-PA with 2.4\% params), often with fewer parameters. Second, the same method applied at FFN always improves over its attention counterpart. For example, LoRA (ffn) improves LoRA (attn) by 1 R-2 points on XSum. We also highlight that prefix tuning does not keep improving when we further increase the capacity, which is also observed in~\citet{li2021prefix}.
% Interestingly, prefix tuning 
% yields better results applying the same modification method to the FFN sub-layer always outperforms its counterpart at attention under various number of parameters (the red markers are above the blue ones).
% For each method we examined, when the modification vector is applied to the attention output, increasing the bottleneck dimension does not always improve the results. In contrast, when the modification occurs at the FFN output, increasing $r$ can consistently improve the performance \cz{add sequential adapter results to support this?}, e.g. the ffn parallel adapter improves the summarization R-2 and MT BLEU by 0.6 and 0.9 points respectively by increasing $r$ from 200 to 1024.
% This demonstrates that the increased capacity of tunable modules can be better utilized when the modules are placed at the FFN sub-layer.
These results suggest that \emph{FFN modification can utilize the added parameters more effectively than attention, no matter what the functional form or composition function is}.
% Second, comparing methods using similar number of parameters at different positions, it's clear to see applying modification at FFN yields better results, for example, LoRA at attention outperforms its counterpart at FFN by 1.02 points on the summarization even with less parameters.
% Thus, to maximize the performance of parameter-efficient tunable modules especially when using larger bottleneck dimensions, FFN is a sweet spot to apply modification vectors. 
We hypothesize that this is because the FFN learns task-specific textual patterns~\citep{geva2020transformer}, while attention learns pairwise positional interactions which do not require large capacity for adapting to new tasks. 
% It's also empirically shown that it's conducive to overparameter the FNN layers in\cz{cite the paper or remove this sentence? we are not overparamterizing actually.}.
% However, as analyzed in \S\ref{sec:multihead} the multi-head advantage of the attention layer allows us to achieve descent performance economically (when $r$ is small).

\paragraph{Is the story different when we use $0.1\%$ parameters?}
In \textsection\ref{sec:reinter-prefix} we reason that prefix tuning is more expressive than adapters (attn), which, however, is not reflected in  Figure~\ref{fig:position}.
We conjecture that this is because multi-head attention is only superior when the parameter budget is small.
% bottleneck dimensions is small, while in Figure~\ref{fig:position} $r$ or $l$ is greater than 200.
% a reasonably size of parameters is practically sufficient for the multi-head prefix tuning, where the rank limitation is not a bottleneck anymore. To validate this hypothesis, we examine the behaviour with a smaller $r$ or $l$ where the rank is a real bottleneck. 
To validate this hypothesis, we compare prefix tuning to parallel adapters when they add $0.1\%$ of the pretrained parameters. 
To ablate the impact of the composition function,
we also report the results of removing the gating in prefix tuning as $\vh + \dvh$. We include the results of the multi-head parallel adapter variant (MH PA) described in \textsection\ref{sec:transfer}. 
% to see whether MH PA is able to improve PA. 
% We perform this analysis on the en-ro dataset.
%
As shown in Table~\ref{tab:res-mh}, the multi-head methods -- prefix tuning and MH PA (attn) -- outperform all others by at least 1.6 BLEU points when using 0.1\% of the parameters. Surprisingly, reducing $l$ from 200 to 30 only causes 0.4 BLEU loss for prefix tuning while PA (attn) loses 1.9 points. The gating composition function in prefix tuning slightly helps the results by 0.3 points. We highlight that the MH parallel adapter improves the single-headed version by 1.6 points, which again verifies the effectiveness of the multi-head formulation.
% Comparing at $r=200$, however, MH PA does not improve much, similarly to prefix tuning. 

Combining the results in Figure~\ref{fig:position} and Table~\ref{tab:res-mh}, we conclude that \emph{modifying head attention shows the best results when the parameter budget is very small, while the FFN can better utilize modifications at larger capacities.} 
% \emph{} indicate that prefix tuning or MH PA admit better performance-parameter tradeoff when $r$ is small, and motivates a better parameter-efficient variant in \textsection\ref{sec:combine} by utilizing the multi-head advantage. 
This suggests that it may be effective to allocate a larger parameter budget to FFN modification instead of treating attention and FFN equally as in~\citet{houlsby2019parameter}.

\subsection{Which Composition Function?}
\label{sec:exp-comp}

We have presented three composition functions in \S\ref{sec:unify}: simple addition (adapter), gated addition (prefix tuning) and scaled addition (LoRA). As it is unnatural to incorporate the exact gated addition into methods whose functional form does not use softmax,
% \gn{``which requires the normalization operation'' is not very clear}, 
we examine the other two by ablating on LoRA and comparing with the proposed scaled parallel adapter (Scaled PA), we constrain modified representation to be FFN since it is generally more effective as shown in \textsection\ref{sec:exp-position}.
\begin{wraptable}{r}{0.45\textwidth}
        \centering
        \vspace{-5pt}
    \caption{Results on XSum when using different composition functions. The modified representation is FFN. The bottleneck dimension $r=512$ for (Scaled) PA and $r=102$ for LoRA. 
    % B-init and L-init represents BERT and LoRA initialization methods respectively.
    }
    \vspace{-5pt}
    \label{tab:composition}
    \resizebox{0.45 \columnwidth}{!}{
    \setlength{\tabcolsep}{2pt}
    \begin{tabular}{lc}
    \toprule
     Method (\# params) & XSum (R-1/2/LSum)  \\
     \midrule
     LoRA (6.1\%), $s$=4 & 44.59/21.31/36.25 \\
     LoRA (6.1\%), $s$=1 & 44.17/20.83/35.74  \\
     PA (6.1\%) & 44.35/20.98/35.98 \\
     \midrule
     Scaled PA (6.1\%), $s$=4 & \textbf{44.85/21.54/36.58} \\
    % Parallel Adapter (LoRA init), $r=512, s=1$ & \textbf{44.48/21.19/36.20} \\
    %  Parallel Adapter (BERT init), trainabled r & 44.48/21.09/36.17 \\
     Scaled PA (6.1\%), trainable $s$ & 44.56/21.31/36.29 \\
     \bottomrule
    \end{tabular}}
    \vspace{-8pt}
\end{wraptable}
Table~\ref{tab:composition} reports the results on XSum. 
We set $r$ as 512 for adapters and 102 for LoRA so that their tuned parameter sizes are the same. We select $s$ based on the R-2 score on the dev set.
% To match the parameters of the adapter with $r=512$ at FFN, we use $r=102$ for LoRA\footnote{There are two weight matrices $W_1\in \sR^{d \times 4d}$ and $W_2\in \sR^{4d \times d}$ in each FFN layer of the pre-trained model. LoRA uses a pair of down-up project matrices for each of the weight matrix, thus to match the number of parameters of adapters that only employ a pair of down-up project matrices, a smaller $r$ is used for LoRA.}.
% First, we notice that LoRA and adapter employ different parameter initialization methods: LoRA uses a random Kaiming uniform~\citep{he2015delving} initialization for $\mW_{\mathrm{down}}$ and zero for $\mW_{\mathrm{up}}$ (LoRA init), while adapters use the same initialization as BERT~\citep{devlin2019bert}.
% \gn{This seems like a detail that could be moved to the appendix maybe. Either way, it shouldn't be mentioned here but perhaps in a footnote when you first introduce these methods?} \cz{this is a pretty important detail and we also reflect it in Tab.~\ref{tab:composition}}. 
We observe that LoRA ($s=4$) performs better than parallel adapter.
% To investigate if the scaled composition is important for LoRA's performance, we replace its initialization with BERT init and set $s$ to be 1. It turns out that LoRA performs similarly as the parallel adapter (LoRA s=1 v.s. PA (BERT init) s=1).
However, the advantage disappears if we remove the scaling by setting $s=1$. 
Through plugging the composition function of LoRA into parallel adapter, the resulted Scaled PA improves the vanilla parallel adapter by 0.56 ROUGE-2 points.
% i.e.~using scaled addition with $s=4$ as the composition function and the LoRA init, and we find that with this technique the performance of PA is substantially improved
% (ROUGE-2 from 20.98 to 21.54).
We also experiment with a learned scalar which does not give better results.
Therefore, we conclude that \emph{the scaling composition function is better than the vanilla additive one while being easily applicable.}

\subsection{An Effective Integration by Transferring Favorable Design Elements}
\label{sec:combine}
\begin{table}[!t]
    \centering
    \small
    \caption{Comparison of various parameter-efficient tuning methods and the proposed variants. 
    % MAM Adapter is one of the proposed methods by mixing and matching the favorable design elements.
    ``$\dagger$'' are results copied from~\citet{lewis-etal-2020-bart} and~\citet{liu-etal-2020-multilingual-denoising}.
    We could not reproduce exactly the same full fine-tuning numbers with the same hyperparameters or even searching them.
    The reason may be the different libraries which the training code is based on -- full fine-tuning is very sensitive to training hyperparameters. For the most performant methods we run with 3 random seeds and report mean and standard deviation.
    % \footnote{Full fine-tuning is very sensitive to the learning rate schedule or batching.}
    }
    \label{tab:combine}
    \vspace{-5pt}
    \resizebox{0.95\columnwidth}{!}{
    \begin{tabular}{lrcc}
    \toprule
     Method & \text{\#} params & XSum (R-1/2/L)  & MT (BLEU)\\
     \midrule
     Full fine-tuning$^{\dagger}$ & 100\% & 45.14/22.27/37.25 & 37.7 \\
     Full fine-tuning (our run) & 100\% & 44.81/21.94/36.83 & 37.3 \\
     \midrule
     Bitfit~\citep{ben2021bitfit} & 0.1\% & 40.64/17.32/32.19 & 26.4 \\
     Prompt tuning~\citep{lester2021power} & 0.1\% & 38.91/15.98/30.83 & 21.0 \\
     Prefix tuning~\citep{li2021prefix}, $l$=200 & 3.6\% & 43.40/20.46/35.51 & 35.6 \\
     Pfeiffer adapter~\citep{pfeiffer2021adapterfusion}, $r$=600 & 7.2\% & 44.03/20.89/35.89$_{\pm\text{.13/.10/.08}}$ & 36.9$_{\pm\text{.1}}$\\
     LoRA (ffn), $r$=102 & 7.2\% & 44.53/21.29/36.28$_{\pm\text{.14/.07/.10}}$ & 36.8$_{\pm\text{.3}}$ \\
     Parallel adapter (PA, ffn), $r$=1024 & 12.3\% & 44.71/21.41/36.41$_{\pm\text{.16/.17/.16}}$ & 37.2$_{\pm\text{.1}}$ \\
     \midrule
     PA (attn, $r$=30) + PA (ffn, $r$=512)  & 6.7\% & 44.29/21.06/36.12$_{\pm\text{.31/.19/.18}}$ & 37.2$_{\pm\text{.1}}$ \\
     Prefix tuning (attn, $l$=30) + LoRA (ffn, $r$=102) & 6.7\% & 44.84/21.71/36.77$_{\pm\text{.07/.05/.03}}$ &  37.0$_{\pm\text{.1}}$ \\
     \midrule
     MAM Adapter (our variant, $l$=30, $r$=512)& 6.7\% & \textbf{45.06/21.90/36.87}$_{\pm\text{.08/.01/.04}}$ & \textbf{37.5}$_{\pm\text{.1}}$ \\
    \bottomrule
    \end{tabular}}
    \vspace{-13pt}
\end{table}
% \begin{table}[!t]
%     \centering
%     \small
%     \caption{Accuracy on dev set of MNLI and SST2. For MNLI we report the overall (matched and mismatched) accuracy. ``Prefix + Scaled PA'' uses smaller bottleneck dimensions to control the number of parameters to be 0.5\%.}
%     \label{tab:classification}
%     % \resizebox{0.8 \columnwidth}{!}{
%     \begin{tabular}{lrr}
%     \toprule
%      Method (\# params) & MNLI & SST2 \\
%      \midrule
%      Full-FT (100\%) & 87.6 & 94.6 \\
%      \midrule
%      BitFit (0.1\%) & 84.7 & 93.7 \\
%      Prefix (0.5\%) & 86.3 & 94.0\\
%      LoRA (0.5\%) & 87.2 & 94.2\\
%      Adapter (0.9\%) & 87.3 & {\bf 94.8}\\
%      \midrule
%      Prefix + Scaled PA (0.5\%) & {\bf 87.4}& 94.2 \\
%     \bottomrule
%     \end{tabular}
%     \vspace{-3mm}
% \end{table}
% So far we have discussed the best design choices of each component of parameter-efficient tuning methods, it is natural to extract the best pieces and instantiate a new and more effective model. 
% As a result of our unified framework of parameter-efficient tuning, 
% Because of our unified framework, we have successfully transferred the best design choices of different methods along the design dimensions, which instantiate more effective models such as parallel adapter (PA), multi-head PA, and scaled PA as demonstrated so far. Here we take a step further, mixing and matching the favorable designs to yield an even better method.
We first highlight three findings in previous sections: (1) Scaled parallel adapter is the best variant to modify FFN; (2) FFN can better utilize modification at larger capacities; and (3) modifying head attentions like prefix tuning can achieve strong performance with only 0.1\% parameters. Inspired by them, we mix and match the favorable designs behind these findings:
% First, as demonstrated in \S\ref{sec:exp:insertion}, prefix tuning is the most effective method when the modification vector is limited to the attention sub-layer, especially with a small bottleneck dimension. 
% Second, we transferred its parallel insertion form to adapter and create a parallel adapter that achieves better performance (\S\ref{sec:exp:insertion}).
% Next, we showed in \S\ref{sec:exp-position} that the FFN is a sweet spot for applying modifications. 
% Finally, we transfered the composition function of scaled addition from LoRA to further improve the parallel adapter.
% Combining the best pieces of each design component gives our new instantiation of a more effective parameter-efficient tuning model.
specifically, we use prefix tuning with a small bottleneck dimension ($l=30$) at the attention sub-layers and allocate more parameter budgets to modify FFN representation using the scaled parallel adapter ($r=512$). Since prefix tuning can be viewed as a form of adapter in our unified framework, we name this variant as \emph{Mix-And-Match adapter (MAM Adapter)}.
In Table~\ref{tab:combine}, we compare MAM adapter with various parameter-efficient tuning methods.
For completeness, we also present results of other combination versions in Table~\ref{tab:combine}: using parallel adapters at both attention and FFN layers and combining prefix tuning (attn) with LoRA (ffn) -- both of these combined versions can improve over their respective prototypes.
However, MAM Adapter achieves the best performance on both tasks and is able to match the results of our full fine-tuning by only updating $6.7\%$ of the pre-trained parameters.
In Table~\ref{tab:res-classification-overview}, we present the results of MAM Adapter on MNLI and SST2 as well, where MAM Adapter achieves comparable results to full fine-tuning by adding only 0.5\% of pretrained parameters.   
% and we can see that all the methods can perform comparably well to the full fine-tuning by only updating 0.5\% of parameters.
% Prompt-tuning prepends $k$ vectors to the input embedding, which significantly increases the memory usage, we choose $K=200$ to fit the memory of an A100 GPU. We have also experimented with larger $k$ (i.e. 300) but it only improves the results marginally.

\section{Discussion}
% hypernetwork of adapter is like learning prompts
We provide a unified framework for several performant parameter-tuning methods, which enables us to instantiate a more effective model that matches the performance of full fine-tuning method through transferring techniques across approaches. 
We hope our work can provide insights and guidance for future research on parameter-efficient tuning.
% and we also suggest these methods to perform experiments on tasks of varying resources and properties to exhibit a more complete view. 

\section*{Ethics Statement}

Our work proposes a method for efficient fine-tuning of pre-trained models, in particular language models.
Pre-trained language models have a wide variety of positive applications, such as the applications to summarization, translation, or language understanding described in our paper.
At the same time, there are a number of ethical concerns with language models in general, including concerns regarding the generation of biased or discriminative text \citep{bordia-bowman-2019-identifying}, the leakage of private information from training data \citep{carlini2020extracting}, and environmental impact of training or tuning them \citep{strubell-etal-2019-energy}.

Our method attempts to train language models making minimal changes to their pre-existing parameters.
While it is an interesting research question whether parameter-efficient fine-tuning methods exacerbate, mitigate, or make little change to issues such as bias or information leakage, to our knowledge no previous work has examined this topic.
It is an interesting avenue for future work.

With respect to environmental impact, the methods proposed in this paper add a small number of extra parameters and components to existing models, and thus they have a nominal negative impact on training and inference time -- for example, the final MAM Adapter needs 100\% - 150\% training time of full fine-tuning in our four benchmarks since parameter-efficient tuning typically needs more epochs to converge; the inference time is roughly the same as the model obtained by full fine-tuning.
On the other hand, as the methods proposed in this paper may obviate the need for full fine-tuning, this may also significantly reduce the cost (in terms of memory/deployed servers) of serving models.
Notably, the great majority of the experimentation done for this paper was performed on a data center powered entirely by renewable energy.

% the following commented for arxiv
\section*{Reproducibility Statement}
In addition to the setup description in \textsection\ref{sec:exp-setup}, we have detailed the complete experiments setup such as batch size, optimizer, learning rates in Appendix~\ref{app:sec:exps}. Besides, we have publicized our source code. These resources should be sufficient to reproduce results of the paper. 

\section*{Acknowledgement}
We thank the anonymous reviewers for their comments. This work was supported in part by the CMU-Portugal MAIA Project, a Baidu PhD Fellowship for Junxian He, and a CMU Presidential Fellowship for Chunting Zhou.

\bibliography{iclr2022_conference}
\bibliographystyle{iclr2022_conference}

\newpage
\appendix
\section{Experiments}
\label{app:sec:exps}
% \gn{I really like Table \ref{tab:design}, and it might make the experiments a bit clearer if you also had similar tables in the appendix for each of the comparisons that you do in each subsection here.}
% \jh{not sure what this means}\cz{me either, maybe ignore}
% In this section, we present additional experimental details .
\subsection{Setups}
\begin{table}[h]
    \centering
    \caption{Dataset Statistics of the four tasks.}
    \begin{tabular}{lccc}
    \toprule
        Dataset & \#train & \#dev & \#test \\
    \midrule
         XSum & 204,045 & 113,332 & 113,334\\
         WMT16 en-ro & 610,320 & 1,999 & 1,999 \\
        MNLI & 392,702 & 9815 & 9832 \\
        SST-2 & 67,349 & 872 & 1,821\\
    \bottomrule
    \end{tabular}
    \label{app:tab:data}
\end{table}
% \jh{
% moved from section 4: 
% }
% \jh{mention dataset number of training examples}
We implement all the parameter-efficient tuning methods using the huggingface transformers library~\citep{wolf-etal-2020-transformers}. 
We use \texttt{BART}$_\text{LARGE}$\citep{lewis-etal-2020-bart} and \texttt{mBART}$_\text{LARGE}$~\citep{liu-etal-2020-multilingual-denoising} (mBART-cc25) for the summarization and machine translation tasks respectively, and we use \texttt{RoBERTa}$_\text{BASE}$~\citep{liu2019roberta} for MNLI and SST2. \texttt{BART}$_\text{LARGE}$ and \texttt{mBART}$_\text{LARGE}$ have the same encoder-decoder architectures. \texttt{mBART}$_\text{LARGE}$ is pre-trained on 25 languages. We use their public checkpoints from the transformers library in experiments. 
For MT and classifications tasks, the max token lengths of training data are set to be 150 and 512 respectively.
For XSum, we set the max length of source articles to be 512 and the max length of the target summary to be 128. The detailed dataset statistics is present in Table~\ref{app:tab:data}. In our summarization experiments, we only use 1600 examples for validation to save time.

While we vary the bottleneck dimension within $\{1, 30, 512, 1024\}$ as mentioned in \textsection\ref{sec:exp-setup}, we test bottleneck dimension 1024 only when the modified representation is FFN, because the training of prefix tuning does not fit into 48GB GPU memory when $l=1024$. While other methods do not have memory issues, we keep the bottleneck dimension of attention modification at most 512 to have a relatively fair comparison with prefix tuning. For LoRA we always tune its scaling hyperparameters $s$ on the dev set.

\subsection{Training and Evaluation}
\begin{table}[!t]
    \centering
    \caption{Training hyperparameters of parameter-efficient tuning methods on the four tasks. lr and ls represents learning rate and label smoothing respectively.}
    \begin{tabular}{lcccccc}
    \toprule
        Tasks & lr & batch size & ls & max grad norm & weight decay & train steps\\
    \midrule
        XSum & 5e-5 & 64 sents & 0.1 & 0.1 & 0.01 & 100K \\
        enro MT & 5e-5 & 16384 tokens & 0.1 & 1.0 & 0.01 & 50K\\
        MNLI/SST2 & 1e-4 & 32 sents & 0 & 1.0 & 0.1 & 10 epochs\\
    \bottomrule
    \end{tabular}
    \label{app:tab:hparams}
\end{table}
We present some training hyperparameters of parameter-efficient tuning methods in Table~\ref{app:tab:hparams}. For all the tasks, we train with the Adam optimizer~\citep{kingma2014adam}, and use a polynomial learning rate scheduler that linearly decays the learning rate throughout training. We set the warm up steps of learning rate to be 0 for both MT and summarization tasks, and for the classification tasks, learning rate is linearly warmed up from 0 for the first 6\% of the total training steps before decay. For full fine-tuning we set these training hyperparameters following~\citet{lewis-etal-2020-bart} (XSum), ~\citet{liu-etal-2020-multilingual-denoising} (en-ro), and~\citep{liu2019roberta} (MNLI and SST2). We also did hyperparameter search in the full fine-tuning case to try to reproduce their results.
We set dropout rate to be 0.1 for all the tasks. 
We use ROUGE-2 and perplexity as the validation metrics for summarization and MT respectively.

For MT and text summarization, we use beam search for decoding and set the number of beams to be 6 and 5 following previous work~\citep{li2021prefix,liu-etal-2020-multilingual-denoising}. The min and max generation lengths for summarization and MT are set to be (10, 60) and (1, 200) respectively.

\subsection{Other Experimental Details}
\paragraph{Prefix Tuning:} Following \citet{li2021prefix},  we reparameterize the prefix vectors by a MLP network which is composed of a small embedding matrix and a large feedforward neural network. This is conducive for learning due to the shared parameters across all layers. 

\paragraph{LoRA:} LoRA and adapter employ different parameter initialization methods: LoRA uses a random Kaiming uniform~\citep{he2015delving} initialization for $\mW_{\mathrm{down}}$ and zero for $\mW_{\mathrm{up}}$ (LoRA init), while adapters use the same initialization as BERT~\citep{devlin2019bert}. We found it beneficial to use the same initialization method as LoRA in scaled PA.

\section{Computation of Tunable Parameters}
\label{app:params}
\begin{figure}[h]
    \vspace{-5mm}
\begin{minipage}{.4\textwidth}
    \begin{table}[H]
        \centering
        \caption{Number of attention or FFN sub-layers in each layer of the pre-trained models.}
        \label{app:tab:num:layers}
    \resizebox{1 \columnwidth}{!}{
        \setlength{\tabcolsep}{2pt}
        \begin{tabular}{lcc}
        \toprule
        % &\texttt{BART}$_{\texttt{LARGE}}$ / \texttt{mBART}$_{\texttt{LARGE}}$  & \texttt{RoBERTa}$_{\texttt{LARGE}}$ \\
        & BART/mBART$_{\mathrm{LARGE}}$ & RoBERTa$_{\mathrm{BASE}}$\\
        \midrule
        $N_{\mathrm{attn}}$ & 3 & 1 \\
        $N_{\mathrm{ffn}}$ & 2 & 1 \\
        \bottomrule
       	\end{tabular}}
        % \vspace{-5mm}
    \end{table}
\end{minipage}
\hfill
\hfill
\begin{minipage}{.57\textwidth}
    \begin{table}[H]
        \centering
        \caption{Number of parameters used at each sub-layer for different methods.}
        \label{app:tab:num:matrices}
        \resizebox{0.95 \columnwidth}{!}{
        \setlength{\tabcolsep}{4pt}
        \begin{tabular}{lcc}
        \toprule
        & $N_{\mathrm{\mW}}^{\mathrm{attn}}$ & $N_{\mathrm{\mW}}^{\mathrm{ffn}}$ \\
        \midrule
        Prefix Tuning & $2ld$ & -- \\
        Adapter variants& $2rd$ & $2rd$ \\
        LoRA & $2\times  2rd=4rd$ & $2 \times (rd + 4dr)=10rd$ \\
        \bottomrule
        \end{tabular}}
        % \vspace{-5mm}
    \end{table}
    \end{minipage}
    \vspace{-3pt}
\end{figure}
We compute the number of tunable parameters based on where the tunable module is inserted into and how it is parameterized.
The pretrained-models for summarization or MT have an encoder-decoder structure and each has $L$ layers, whereas \texttt{RoBERTa$_\text{BASE}$} for classification tasks only has $L$ encoder layers.
To simplify the computation of tunable parameters, we compute the sum of parameter used in one encoder layer and one decoder layer as the parameter overhead of one single layer of the pre-trained encoder-decoder model.
Each layer has $N_{\mathrm{attn}}$ sub-layers and $N_{\mathrm{ffn}}$ sub-layers.
For the encoder-decoder models, $N_{\mathrm{attn}}=3$: the encoder self-attention, the decoder self-attention and the decoder cross-attention. For the classification tasks, \texttt{RoBERTa$_\text{BASE}$} only has the encoder self-attention, thus $N_{\mathrm{attn}}=1$.
We present the number of attention and ffn sub-layers for different pre-trained models in Table~\ref{app:tab:num:matrices}.
For modifications applied at the attention sub-layers, the number of tunable parameters is computed by $|\Theta|_{\mathrm{attn}} = N_{\mathrm{\mW}}^{\mathrm{attn}} \times N_{\mathrm{attn}} \times L$, where $N_{\mathrm{\mW}}^{\mathrm{attn}}$ denotes the number of parameters ($\mW_{\mathrm{down}}$ or $\mW_{\mathrm{up}}$) used for one attention sub-layer. Similarly, the number of tunable parameters for the FFN sub-layers is computed by $|\Theta|_{\mathrm{ffn}} = N_{\mathrm{\mW}}^{\mathrm{ffn}} \times N_{\mathrm{ffn}} \times L$. 
In Table~\ref{app:tab:num:matrices}, we show the number of parameters for one sub-layer. As we have explained in \S\ref{sec:exp-position}, LoRA approximates the update of each weight matrix with a pair of $\mW_{\mathrm{down}}$ and $\mW_{\mathrm{up}}$, thus LoRA typically uses more parameters with the same $r$ as other methods. 
Finally, the total number of tunable parameters for prefix tuning, adapter variants and LoRA is $|\Theta|=|\Theta|_{\mathrm{attn}}+|\Theta|_{\mathrm{ffn}}$ as applicable. 
Prompt tuning prepends $l$ tunable vectors at the input layer and uses $l\times d$ number of parameters.
Using MBART/BART as an example, we present the number of parameters used by several representative methods throughout our paper in Table~\ref{app:tab:examples:params}, where adapter variants include sequential adapter, parallel adapter, scaled adapter and multi-head adapter.

\begin{table}[h]
    \centering
    \caption{Number of tunable parameters of various parameter-efficient tuning methods with BART/MBART models ($L=12$) as an example.}
    \label{app:tab:examples:params}
    \begin{tabular}{lc}
    \toprule
    Method & number of parameters \\
    \midrule
        Prompt Tuning & $l \times d$ \\
        Prefix Tuning (attn) & $2ld \times 3 \times 12$ \\
        Adapter variants (attn) & $2rd \times 3 \times 12$ \\
        Adapter variants (ffn) & $2rd \times 2 \times 12$\\
        LoRA (attn) & $4rd \times 3 \times 12$ \\
        LoRA (ffn) & $ 10 rd \times 2 \times 12$\\
        MAM Adapter (our proposed model) & $2ld \times 3 \times 12 + 2rd \times 2 \times 12$\\
        \bottomrule
    \end{tabular}
\end{table}

\section{Full Results on Different Bottleneck Dimensions}
\label{app:results}
\begin{table}[!t]
\small
    \centering
    \caption{Performance on the test sets of abstractive summarization (XSum) and WMT EN-RO translation.}
    \label{tab:position}
    % \resizebox{0.8 \columnwidth}{!}{
    \begin{tabular}{lrcr}
    \toprule
     Method & \text{\#} params (\%) & XSum (R-1/2/L) & MT BLEU \\
     \midrule
    %  full fine-tuning & 100 & 44.81/21.94/36.83 & 37.3 \\
    %  \midrule
     \multicolumn{4}{c}{Modified Representation: attention} \\
     Prefix Tuning, $r=200$ & 3.6 & 43.40/20.46/35.51 & 35.6 \\
     Prefix Tuning, $r=512$ & 9.2 & 43.29/20.40/35.37 & 35.1 \\
     LoRA, $r=200$ & 7.2 & 43.09/20.29/35.37 & 36.2 \\
     Sequential Adapter, $r=200$ & 3.6 & 42.01/19.30/34.40 & 35.3 \\
     Sequential Adapter, $r=512$ & 9.2 & 41.05/18.87/33.71 & 34.7 \\
     Parallel Adapter, $r=200$ & 3.6 & 43.58/20.31/35.34 & 35.6 \\
     Parallel Adapter, $r=512$ & 9.2 & 43.99/20.83/35.77 & 36.2 \\
     \midrule
     \multicolumn{4}{c}{Modified Representation: FFN}\\
     LoRA, $r=102$ & 6.1 & 44.59/21.31/36.25 & 36.5 \\
     Sequential Adapter, $r=200$ & 2.4 & 43.21/19.98/35.08 & 35.6\\
     Sequential Adapter, $r=512$ & 6.1 & 43.72/20.75/35.64 & 36.3 \\
     Sequential Adapter, $r=1024$ & 12.3 & 43.95/21.00/35.90 & 36.7 \\
     Parallel Adapter, $r=200$ & 2.4 & 43.93/20.66/35.63 & 36.4 \\
     Parallel Adapter, $r=512$ & 6.1 & 44.35/20.98/35.98 & 37.1 \\
     Parallel Adapter, $r=1024$ & 12.3 & 44.53/21.24/36.23 & 37.3 \\
    \bottomrule
    \end{tabular}
    % \vspace{-5mm}
\end{table}

\end{document}